\pdfoutput=1
\documentclass[11pt]{article}

\usepackage[preprint]{acl}
\usepackage{colortbl}

\usepackage{times}
\usepackage{latexsym}
\usepackage{booktabs}
\usepackage{tikz}
\usepackage{pgfplots}
\usepackage{amssymb}
\usepackage{amsmath}
\usepackage{enumitem}
\usepackage{pifont}
\usetikzlibrary{positioning, shapes.multipart, fit, calc, shapes.geometric, shapes.misc}
\usepackage{subcaption}
\usepackage{multirow}
\usepackage[T1]{fontenc}
\usepackage[utf8]{inputenc}

\usepackage{microtype}

\usepackage{inconsolata}

\usepackage{graphicx}
\usepackage{makecell}
\usepackage{pgfplotstable}
\usepackage{filecontents}

\title{Asymmetric Conflict and Synergy in Post-training for LLM-based Multilingual Machine Translation}

\author{
 \textbf{Tong Zheng\textsuperscript{1}},
 \textbf{Yan Wen\textsuperscript{1}},
 \textbf{Huiwen Bao\textsuperscript{3}},
 \textbf{Junfeng Guo\textsuperscript{1,2}},
 \textbf{Heng Huang\textsuperscript{1,2}}
\\
\\
 \textsuperscript{1}Department of Computer Science, University of Maryland College Park\\
  \textsuperscript{2}Institute of Health Computing, University of Maryland College Park\\
 \textsuperscript{3}City University of Hong Kong
\\
 \small{
   \href{mailto:zhengtong12356@gmail.com}{\{tzheng24, gjf2023, heng\}@umd.edu},
   \href{mailto:goodbaohuiwen@gmail.com}{goodbaohuiwen@gmail.com},
 }
}

\begin{document}
\maketitle
\begin{abstract}
The emergence of Large Language Models (LLMs) has advanced the multilingual machine translation (MMT), yet the \textit{Curse of Multilinguality} (CoM) remains a major challenge. Existing work in LLM-based MMT typically mitigates this issue via scaling up training and computation budget, which raises a critical question: \textit{Is scaling up the training and computation budget truly necessary for high-quality MMT, or can a deeper understanding of CoM provide a more efficient solution?} To explore this problem, we analyze the linguistic conflicts and synergy, the underlying mechanism of CoM during post-training phase. We identify an \textit{asymmetric phenomenon in linguistic conflicts and synergy}: the dominance of conflicts and synergy varies in different translation directions, leading to sub-optimal adaptation in existing post-training methods. We further find that a significant bottleneck in MMT appears to lie in post-training rather than multilingual pre-training, suggesting the need for more effective adaptation strategies. Building on these new insights, we propose a direction-aware training approach, combined with group-wise model merging, to address asymmetry in linguistic conflicts and synergy explicitly. Leveraging this strategy, our method fine-tunes X-ALMA-13B-Pretrain—trained only with multilingual pre-training—achieving comparable performance to XALMA-13B (only SFT) while using only 20B pretraining tokens and 17B parameters—5.5× fewer pretraining-tokens and 1.7x fewer model size—with just 0.85 COMET drop on Flores-200 testsets of 50 languages.

\end{abstract}

\section{Introduction}
Large language models (LLMs) have shown remarkable general capabilities~\cite{brown2020language,wei2022chain,dubey2024llama} and have advanced multilingual machine translation~\cite{xu2024a,yang2023bigtranslate,alves2024tower}. For example, Aya-101~\cite{aryabumi2024aya} expands support to 101 languages and achieves strong performance in multilingual machine translation, while LLaMAX~\cite{lu2024llamax} further pushes performance beyond 100 languages. The common practice behind these successes is the large-scale pretraining, which typically involves monolingual pretraining~\footnote{We also refer to this as multilingual pretraining, where data from all languages are mixed during the pretraining process.}, parallel pretraining, or both—followed by a small-scale, high-quality post-training phase. However, as LLMs scale to more languages, they suffer from the issue of \textit{Curse of Multilinguality (CoM)} ~\cite{conneau2019unsupervised}, which degrades the translation performance. 

\usetikzlibrary{shapes.geometric} 
\pgfdeclareplotmark{mystar}{
    \node[star,star point ratio=2.25,minimum size=6pt,
          inner sep=0pt,draw=black,solid,fill=red] {};
}
\definecolor{tiffanyblue}{RGB}{129,216,208}
\definecolor{bangdiblue}{RGB}{0,149,182}
\definecolor{kleinblue}{RGB}{0,47,167}
\definecolor{kabuliblue}{RGB}{26,85,153}
\definecolor{purple}{RGB}{138,43,226}
\definecolor{upink}{RGB}{255,150,128}
\begin{figure}[t!]
\centering
\begin{tikzpicture}
  \pgfplotsset{set layers}
      \begin{axis}[
	 at={(0,0)},
      ymajorgrids,
      xmajorgrids,
      grid style=dashed,
      width=0.47*\textwidth,
      height=0.34*\textwidth,
      legend style={at={(0.23,0.08)}, anchor=south west},
      xlabel={\small{Pre-Training Tokens (B)}},
      ylabel={\small{COMET-22}},
      ylabel style={font=\bfseries,yshift=-1em, xshift=0em},xlabel style={font=\bfseries,xshift=0em,yshift=0.0em},
      yticklabel style={/pgf/number format/precision=0,/pgf/number format/fixed zerofill},
      ymin=83,ymax=90, ytick={83, 85, 87, 89},
      xmin=0,xmax=130,xtick={20, 50, 80, 110},
      legend style={yshift=-6pt,xshift=-2em, legend plot pos=right,font={\footnotesize},cells={anchor=west}}
      ]
    \addplot[kleinblue!30,mark=*,mark size=14pt,thick,mark options={fill=kleinblue!30,draw=kleinblue!30,line width=1.0pt}] coordinates { (110,88.58)
    };
    \node at (axis cs:100,89.7) [font=\bfseries\sffamily\tiny] {X-ALMA-13B (Only SFT)};

    \addplot[orange!70,mark=*,mark size=8.2pt,thick,mark options={fill=orange!70, draw=orange!70,line width=1.0pt}] coordinates { (20,87.81)
    };
    \node at (axis cs:25,88.5) [font=\bfseries\sffamily\tiny] {X-ALMA-DAT (Ours)};

    \addplot[tiffanyblue!70,mark=*,mark size=6.8pt,thick,mark options={fill=tiffanyblue!70, draw=tiffanyblue!70,line width=1.0pt}] coordinates { (20,87.61)
      };
      
    \addplot[purple!30,mark=*,mark size=5.5pt,thick,mark options={fill=purple!30,line width=1.0pt}] coordinates { (66,85.00)
      };
    \node at (axis cs:66,84.2) [font=\bfseries\sffamily\tiny] {LLaMAX};

    \addplot[only marks,red!50,mark=mystar,mark size=5pt,thick,mark options={fill=red!50, draw=red!50,line width=1.0pt}, scale=2] coordinates {(10, 89.2)
    };
    \node at (axis cs:21,89.50) [orange, font=\bfseries\sffamily\small] {Ideal};

    \draw[dashed, ->, ultra thick] 
    (axis cs:96, 88.58) -- (axis cs:29, 87.81);

    \node[red, align=center, font=\bfseries\sffamily\tiny, fill=white, opacity=0.8, text opacity=1, rounded corners, rotate=8] 
   at (axis cs:73, 87.6) 
   {5.5× Pre-Training Cost ↓ \\ 1.7×-2× Model Size ↓};
   \node at (axis cs:26,86.8) [font=\bfseries\sffamily\tiny] {X-ALMA-DATM (Ours)};
    
      \end{axis}

  \end{tikzpicture}
\vskip -0.1in
    \caption{The relationship between pre-training cost, model capacity and translation performance. We evaluate performance on the Flores-200 test sets across 50 languages. \textbf{The size of circle denotes model capacity.}}
    \label{fig:curse_of_multilinguality}
\end{figure}
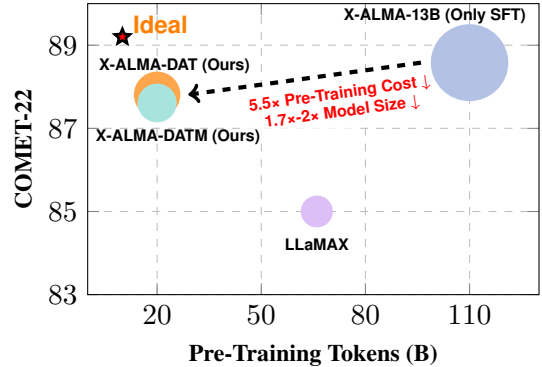

Understanding and mitigating CoM is not new in the MMT literature. In traditional MMT, existing research has identified critical factors such as resource imbalances, limited model capacity, linguistic similarity, and complex interactions between language pairs, particularly for low-resource languages~\cite{arivazhagan2019massively,aharoni-etal-2019-massively,shaham-etal-2023-causes,meng-monz-2024-disentangling}, and proposed solutions including language-specific modules~\cite{fan2021beyond,zhao2024sparse,xu-etal-2023-condensing}, vocabulary optimization~\cite{han2024adapters}, data sampling techniques~\cite{wang-etal-2020-balancing,wang-neubig-2019-target,lin-etal-2019-choosing}, and continual learning approach~\cite{liu-etal-2023-continual}. 
Based on these studies, recent LLM-based MMT research focuses on designing increasingly complex training pipelines and modular architectures. For instance, \citet{xu2024x} proposed a five-stage training pipeline incorporating language-specific modules. However, existing analyses primarily focus on the encoder-decoder paradigm, while current LLM-based approaches heavily rely on scaling up model capacity and computational resources, making them prohibitively expensive. This raises a critical question: \textit{Is scaling up the training and computation budget truly necessary for high-quality MMT, or can a deeper understanding of CoM in LLM-based MMT provides a more efficient solution?}

In this work, we systematically investigate linguistic conflicts and synergy during post-training phase. We conduct extensive experiments with different settings: across 5 to 50 languages, three pretrained LLMs - ALMA-7B-Pretrain, ALMA-13B-Pretrain and X-ALMA-13B-Pretrain, three distinct post-training strategies - multilingual training, group multilingual training, and separate training. We observe a consistent pattern: asymmetry in linguistic conflicts and synergy (Figure \ref{fig:Asymmetric_Conflict_and_Synergy}, Appendix \ref{app:Asymmetry_in_Linguistic_Conflicts_and_Synergy_in_terms_of_SacreBLEU}and \ref{app:more_experiments_Asymmetry_in_Linguistic_Conflicts_and_Synergy} ). For example, in multilingual training, XX$\rightarrow$En translation directions experience significant linguistic conflicts, leading to performance degradation, whereas En$\rightarrow$XX translations benefit from linguistic synergy, where XX denotes 49 different languages other than English. We further show this asymmetric phenomenon cannot be easily mitigated through existing training approaches, such as group multilingual training (Table \ref{tab:inefficiency_existing_approach}). \textbf{This finding illustrates the need to develop a direction-aware training strategy for optimal post-training.}

Another key finding of our work is that a simple multilingual pre-training stage can be sufficient to equip foundation models with ideal multilingual capabilities, whereas the bottleneck lies in the post-training stage (dotted lines in Figure \ref{fig:Asymmetric_Conflict_and_Synergy} (g-i)). 
Motivated by these findings, we propose a novel \textit{Direction-Aware Training} (DAT) approach and build an efficient MMT starting from a relatively efficient base model, the X-ALMA-13B-Pretrain—utilizing only simple multilingual pre-training on 20 billion tokens. Our approach fully leverages the interactive characteristics of different language directions to reduce conflicts while maximizing synergy. We also present a scalable version of the approach, named DATM, which utilizes model merging to further enhance efficiency with only negligible performance degradation.

Through comprehensive evaluations on Flores-200 and WMT23 Benchmark, we demonstrate the effectiveness of our approach. Notably, as shown in Figure \ref{fig:curse_of_multilinguality}, compared to X-ALMA (Only SFT)~\cite{xu2024x}, our model X-ALMA-13B-DAT maintains comparable performance while having two advantages: 1) utilizing a simple and efficient training recipe - starting from base models with fewer pre-training tokens and employing a post-training stage. 2) parameter-efficient - we consume 1.7x fewer parameters compared to X-ALMA (Only SFT). These results demonstrate that simple pre-training combined with dedicated post-training can also achieve good multilingual performance.

\section{Experimental Settings}
In this section, we introduce the basic experimental settings used in Section \ref{sec:Asymmetry_in_Linguistic_Conflicts_and_Synergy} and Section \ref{sec:dtam_approach}. 
\label{sec: settings}
\subsection{Datasets}
\label{subsec:datasets}

We use the high-quality parallel dataset curated by \citet{xu2024x}, covering fifty languages across low-, medium-, and high-resource categories. Following \cite{xu2024x}, these languages are grouped into eight linguistic groups based on linguistic similarity and a balanced number of languages. Details are provided in Section \ref{app:detailed_experimental_setups} in Appendix. 
The dataset primarily consists of samples from the Flores-200 development set and NTREX~\cite{barrault2019findings}. For languages in both Flores-200 and WMT’15-22, corresponding test sets are incorporated, yielding an average of 4K examples per language. For evaluation, we use Flores-200 and WMT23 benchmarks to assess performance. 

\subsection{Models}
\label{subsec:models}
We select three representative fully open multilingual LLMs for our study: ALMA-Pretrain~\cite{xu2024a} (7B–13B parameters) and X-ALMA-Pretrain~\cite{xu2024x} (13B parameters). The ALMA-Pretrain models were pre-trained on 12B or 20B tokens across six languages, while X-ALMA-Pretrain underwent continued pre-training on 20B tokens from 50 languages, both based on LLaMA-2.
We exclude other state-of-the-art multilingual models for two key reasons: (1) their pre-trained checkpoints are unavailable, as in the case of Aya-series~\cite{aryabumi2024aya} and BigTrans~\cite{yang2023bigtranslate}; or (2) they exhibit suboptimal multilingual performance in certain languages as shown in \citet{xu2024x,cui2025multilingual}, such as LLaMA-3~\cite{dubey2024llama}. 

\subsection{Training}
\paragraph{Fine-tuning Strategies}
We employ three distinct training strategies for fine-tuning the models: Multilingual Training, Separate Training, and Group Multilingual Training.
\begin{itemize}[leftmargin=*, itemsep=0pt, parsep=0pt, topsep=0pt, partopsep=0pt]
    \item \textbf{Multilingual Training}~\cite{tang2020multilingual}: 
    This is typically achieved by mixing data from all languages and using it to fine-tune the model. The resulting model is a single model that possesses shared representations across all languages.
    \item \textbf{Group Multilingual Training}~\cite{xu2024x}: 
    We group the languages and then apply multilingual training within each group, resulting in multiple models, each for its respective languages.
    \item \textbf{Separate Training}: 
    Separate tuning involves training a distinct model for each language without considering linguistic synergies or conflicts.
\end{itemize}
\paragraph{Training Configurations}
In this work, all models are trained with a learning rate of 2e-3 using an inverse square root scheduler, a weight decay of 0.01, and a warmup ratio of 0.01. The total batch size is set to 128. Fine-tuning is conducted for 1 epoch, with both \texttt{max\_new\_tokens} and \texttt{max\_source\_length} set to 512. Additionally, FP16 precision training is enabled to optimize performance and efficiency. All models are trained on  4 NVIDIA H100 with LoRA~\cite{hu2022lora} as \citet{xu2024a} has shown a negligible performance gap between LoRA tuning and full fine-tuning.

\subsection{Evaluation}
We set the number of beams to 5 and both \texttt{max\_new\_tokens} and \texttt{max\_source\_tokens} to 512. We evaluate performance mainly using COMET-22~\cite{rei-etal-2022-comet} and SacreBLEU~\cite{post-2018-call}.
\definecolor{tiffanyblue}{RGB}{129,216,208}
\definecolor{bangdiblue}{RGB}{0,149,182}
\definecolor{kleinblue}{RGB}{0,47,167}
\definecolor{kabuliblue}{RGB}{26,85,153}
\definecolor{purple}{RGB}{138,43,226}
\usepgfplotslibrary{groupplots}
\usepgfplotslibrary{fillbetween} 
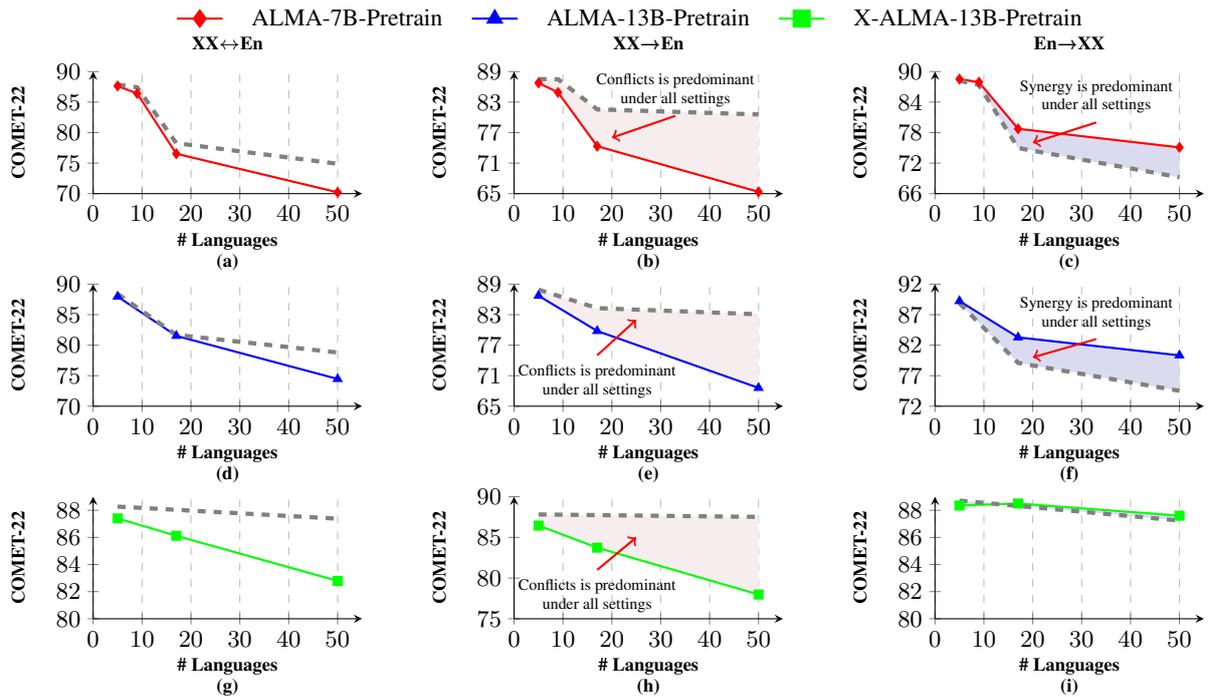
\begin{figure*}[t!]
  \centering
  \begin{tikzpicture}
    \pgfplotsset{set layers}
    \scriptsize{
    % 全局图例单独绘制
    \begin{axis}[
    at={(2,1.4)},
    width=0.5\textwidth,
      height=0.2\textwidth,
      hide axis,
      xmin=0, xmax=0.5, ymin=0, ymax=0.5,
      legend style={
        draw=none,
        fill=none,
        font=\small,
        column sep=0.3cm,
        at={(1.1,1.6)},
        anchor=north
      },
      legend columns=3
    ]
      \addlegendimage{red, mark=diamond*, mark size=3.0pt, thick, mark options={solid, fill=red}}
      \addlegendentry{ALMA-7B-Pretrain}
      \addlegendimage{blue, mark=triangle*, mark size=3.0pt, thick, mark options={solid, fill=blue}}
      \addlegendentry{ALMA-13B-Pretrain}
      \addlegendimage{green, mark=square*, mark size=3.0pt, thick, mark options={solid, fill=green}}
      \addlegendentry{X-ALMA-13B-Pretrain}
    \end{axis}
    }
    \scriptsize{
    \begin{groupplot}[
      group style={group size=3 by 3, horizontal sep=2cm, vertical sep=1.2cm},
      xmajorgrids,
      grid style={dashed, gray!50},
      width=0.32\textwidth,
      height=0.2\textwidth,
      xlabel={\# Languages},
      ylabel={COMET-22},
      xlabel style={font=\bfseries, yshift=0.5em},
      ylabel style={font=\bfseries, yshift=-1.0em},
      yticklabel style={font=\small},
      xticklabel style={font=\small},
      title style={font=\bfseries},
      xmin=0, xmax=55, 
      xtick={0, 10, 20, 30, 40, 50},
        axis x line=bottom,
        axis y line=left,
        % axis y line*=left, % 左Y轴
    ]
    % 第一行：Model 1
    \nextgroupplot[title={XX$\leftrightarrow$En}, xlabel={\makecell{\# Languages \\ (a) } },ymin=70, ymax=90, ytick={70, 75, 80, 85,90}]
    \addplot[red, mark=diamond*, mark size=1.5pt, thick, mark options={solid, fill=red}] 
      coordinates {(5, 87.64) (9, 86.41) (17, 76.54) (50, 70.22)};
    \addplot[ultra thick, dashed, gray] 
      coordinates {(5, 87.86) (9, 87.42) (17, 78.27) (50, 74.89)};
      
    \nextgroupplot[title={XX→En}, xlabel={\makecell{\# Languages \\ (b) } }, ymin=65, ymax=89, ytick={65, 71, 77, 83, 89}]
    \addplot[name path=mt_b,red, mark=diamond*, mark size=1.5pt, thick, mark options={solid, fill=red}] 
      coordinates {(5, 86.76) (9, 84.89) (17, 74.33) (50, 65.35)};
    \addplot[name path=separate_b,ultra thick, dashed, gray] 
      coordinates {(5, 87.53) (9, 87.48) (17, 81.57) (50, 80.57)};
    
    \addplot [
        fill=red!30!gray!10
    ] fill between [
        of=mt_b and separate_b,
    ];

    \draw[->, thick, color=red] (axis cs:33,80.3) node[above]{\tiny \textcolor{black}{\makecell{Conflicts is predominant\\ under all settings}}} -- (axis cs:20,76);

    \nextgroupplot[
        title={En→XX}, 
        xlabel={\makecell{\# Languages \\ (c) } }, 
        % 左 Y 轴（COMET-22）
        ylabel={COMET-22}, 
        ymin=66, ymax=90, 
        ytick={66, 72, 78, 84, 90},
    ]
    \addplot[name path=mt_c, red, mark=diamond*, mark size=1.5pt, thick, mark options={solid, fill=red}] 
      coordinates {(5, 88.52) (9, 87.92) (17, 78.75) (50, 75.09)};
    \addplot[name path=separate_c,ultra thick, dashed, gray] 
      coordinates {(5, 88.19) (9, 87.35) (17, 74.97) (50, 69.21)};

    \addplot [
        fill=blue!40!gray!20
    ] fill between [
        of=mt_c and separate_c,
    ];

    \draw[->, thick, color=red] (axis cs:33,80) node[above]{\tiny \textcolor{black}{\makecell{Synergy is predominant\\ under all settings}}} -- (axis cs:20,76)
         ;

    % 第二行：Model 2
    \nextgroupplot[xlabel={\makecell{\# Languages \\ (d) } }, ymin=70, ymax=90, ytick={70, 75, 80, 85, 90}]
    \addplot[blue, mark=triangle*, mark size=1.5pt, thick, mark options={solid, fill=blue}] 
      coordinates {(5, 87.97) (17, 81.53) (50, 74.46)};
    
    % \addplot[ultra thick, dashed, gray] 
    %   coordinates {(5, 88.40) (9, 88.57) (17, 81.692) (50, 78.80)};

    \addplot[ultra thick, dashed, gray] 
      coordinates {(5, 88.40) (17, 81.692) (50, 78.80)};
    
    \nextgroupplot[xlabel={\makecell{\# Languages \\ (e) } }, ymin=65, ymax=89, ytick={65, 71, 77, 83, 89}]
    \addplot[name path=mt_e, blue, mark=triangle*, mark size=1.5pt, thick, mark options={solid, fill=blue}] 
      coordinates {(5, 86.72) (17, 79.79) (50, 68.60)};

    % \addplot[ultra thick, dashed, gray] 
    %   coordinates {(5, 87.92) (9, 88.068) (17, 84.289) (50, 83.079)};
      \addplot[name path=separate_e, ultra thick, dashed, gray] 
      coordinates {(5, 87.92) (17, 84.289) (50, 83.079)};

     \addplot [
        fill=red!30!gray!10
    ] fill between [
        of=mt_e and separate_e,
    ];

    \draw[->, thick, color=red] (axis cs:17,75) node[below]{\tiny \textcolor{black}{\makecell{Conflicts is predominant\\ under all settings}}} -- (axis cs:25,82)
         ;
     
    \nextgroupplot[xlabel={\makecell{\# Languages \\ (f) } },ymin=72, ymax=92, ytick={72, 77, 82, 87, 92}]
    \addplot[name path=mt_f,blue, mark=triangle*, mark size=1.5pt, thick, mark options={solid, fill=blue}] 
      coordinates {(5, 89.23) (17, 83.28) (50, 80.33)};

     % \addplot[ultra thick, dashed, gray] 
     %  coordinates {(5, 88.882) (9, 89.071) (17, 79.095) (50, 74.52)};

      \addplot[name path=separate_f,ultra thick, dashed, gray] 
      coordinates {(5, 88.882)  (17, 79.095) (50, 74.52)};

      \addplot [
        fill=blue!40!gray!20
    ] fill between [
        of=mt_f and separate_f,
    ];

    \draw[->, thick, color=red] (axis cs:33,83) node[above]{\tiny \textcolor{black}{\makecell{Synergy is predominant\\ under all settings}}} -- (axis cs:20,80)
         ;

    % 第三行：Model 3
    \nextgroupplot[xlabel={\makecell{\# Languages \\ (g) } }, ymin=80, ymax=89, ytick={80, 82, 84, 86, 88}]
    \addplot[green, mark=square*, mark size=1.5pt, thick, mark options={solid, fill=green}] 
      coordinates {(5, 87.40) (17, 86.12) (50, 82.79)};
    \addplot[ultra thick, dashed, gray]  coordinates {(5, 88.27) (50, 87.38)};

    \nextgroupplot[xlabel={\makecell{\# Languages \\ (h) } }, ymin=75, ymax=90, ytick={75, 80, 85, 90}]
    \addplot[name path=mt_h,green, mark=square*, mark size=1.5pt, thick, mark options={solid, fill=green}] 
      coordinates {(5, 86.45) (17, 83.74) (50, 77.98)};
    \addplot[name path=separate_h, ultra thick, dashed, gray] 
      coordinates {(5, 87.82) (50, 87.52)};
    % \node at (0.5,-10.25) {(c) XX-En};

    \addplot [
        fill=red!30!gray!10
    ] fill between [
        of=mt_h and separate_h,
    ];

    \draw[->, thick, color=red] (axis cs:17,81) node[below]{\tiny \textcolor{black}{\makecell{Conflicts is predominant\\ under all settings}}} -- (axis cs:25,85)
         ;
    
    \nextgroupplot[xlabel={\makecell{\# Languages \\ (i) } },ymin=80, ymax=89, ytick={80,82, 84, 86, 88}]
    \addplot[name path=mt_i,green, mark=square*, mark size=1.5pt, thick, mark options={solid, fill=green}] 
      coordinates {(5, 88.35) (17, 88.50) (50, 87.60)};
      \addplot[name path=separate_i, ultra thick, dashed, gray]  coordinates {(5, 88.71) (50, 87.24)};

    \addplot [
        fill=blue!40!gray!20
    ] fill between [
        of=mt_i and separate_i,
    ];

    \end{groupplot}
    }
  \end{tikzpicture}
  \vskip -0.1in
  \caption{Performance of different models trained on varying numbers of languages. The dotted line represents the performance of separately trained models, serving as a reference point where no language conflicts or synergies occur. Two key findings emerge: (1) Asymmetry in Linguistic Conflicts and Synergy (Figure a–i), highlighting the uneven impact of multilingual training across language pairs; and (2) The Bottleneck of Multilinguality in Post-Training (Figure g–i): While multilingual pre-training provides a solid foundation for handling multiple languages, the multilingual training phase can lead to the CoM.}
  \label{fig:Asymmetric_Conflict_and_Synergy}
\end{figure*}
\usetikzlibrary{patterns}
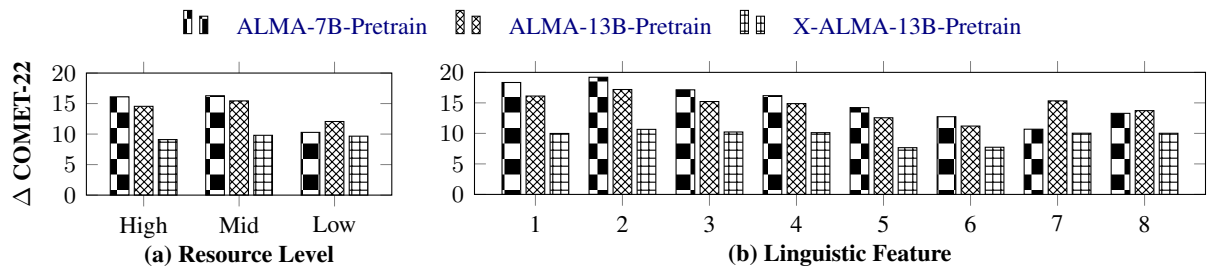
\begin{figure*}[t!]
  \centering
  % 第一行：两图并排
  \ref{sharedlegend}
  \begin{tikzpicture}
    \begin{axis}[
      ybar,
      bar width=7pt,  % 让 bar 变宽以填充空隙
      width=0.35\textwidth,
      height=0.20\textwidth,
      xlabel={\small{(a) Resource Level}},
      ylabel={\small{$\Delta$ COMET-22}},
      xlabel style={font=\bfseries},
      ylabel style={font=\bfseries,yshift=-1em},
      symbolic x coords={High, Mid, Low},
      xtick=data,
      xticklabel style={font=\small},
      yticklabel style={font=\small},
      ymin=0, ymax=20,
      % nodes near coords,
      % nodes near coords style={font=\tiny},
      enlarge x limits=0.3,  % 确保不同类别间有间距
      legend style={
        draw=none,
        fill=none,
        font=\small,
        column sep=0.3cm,
        at={(1,3.7)},
        anchor=north
      },
      legend columns=3,
      legend to name=sharedlegend
    ]
      \addplot[fill=red!80, pattern=checkerboard] coordinates {(High, 16.106) (Mid, 16.28) (Low, 10.279)};
      \addlegendentry{ALMA-7B-Pretrain}
      \addplot[fill=blue!80, pattern=crosshatch] coordinates {(High, 14.548) (Mid, 15.438) (Low, 12.061)};
      \addlegendentry{ALMA-13B-Pretrain}
      \addplot[fill=green!80, pattern=grid] coordinates {(High, 9.1) (Mid, 9.8) (Low, 9.657)};
      \addlegendentry{X-ALMA-13B-Pretrain}
    \end{axis}
  \end{tikzpicture}
  \hspace{0.3cm}
  % 第二行：单图居中
  \vspace{0.2cm}
  \begin{tikzpicture}
    \begin{axis}[
      ybar,
      bar width=7pt,  
      width=0.7\textwidth,
      height=0.20\textwidth,
      xlabel={\small{(b) Linguistic Feature}},
      xlabel style={font=\bfseries},
      symbolic x coords={1, 2, 3, 4, 5, 6, 7, 8},
      xtick=data,
      xticklabel style={font=\small},
      yticklabel style={font=\small},
      ymin=0, ymax=20,
      % nodes near coords,
      % nodes near coords style={font=\tiny},
      enlarge x limits=0.1,  
    ]
      \addplot[fill=red!80, pattern=checkerboard] coordinates {
        (1, 18.34) (2, 19.202) (3, 17.138) 
        (4, 16.182) (5, 14.224) 
        (6, 12.734) (7, 10.686) (8, 13.285)};
      \addplot[fill=blue!80, pattern=crosshatch] coordinates {
        (1, 16.12) (2, 17.19) (3, 15.22) 
        (4, 14.88) (5, 12.56) 
        (6, 11.2) (7, 15.33) (8, 13.73)};
      \addplot[fill=green!80, pattern=grid] coordinates {
        (1, 9.99) (2, 10.66) (3, 10.23) 
        (4, 10.12) (5, 7.66) 
        (6, 7.73) (7, 10.03) (8, 10.02)};
    \end{axis}
  \end{tikzpicture}
  \vskip -0.2in
  \caption{$\Delta$ COMET-22 between separate training and multilingual training in XX → En translation, grouped by resource level and linguistic features. The magnitude of $\Delta$ COMET-22 denotes the intensity of linguistic conflicts.}
  \label{fig:performance_gap_vs_language_group_and_resource_level}
\end{figure*}
\section{The Phenomenon: \textit{Asymmetry in Linguistic Conflicts and Synergy}}
\label{sec:Asymmetry_in_Linguistic_Conflicts_and_Synergy}
In this section, we investigate the phenomenon of \textit{Asymmetry in Linguistic Conflicts and Synergy} in LLM-based MMT. We begin by illustrating the phenomenon (Section \ref{subsec:Asymmetry_in_Linguistic_Conflicts_and_Synergy}) and analyzing its distribution across two essential factors: language resources and groups (Section \ref{subsec:Asymmetry_in_Linguistic_Conflicts_and_Synergy_Across_Language_groups_and_resources}). Finally, we show how this phenomenon poses challenges to existing post-training strategies (Section \ref{subsec:Challenges_by_Asymmetry_in_Linguistic_Conflicts_and_Synergy}).

\subsection{Asymmetry in Linguistic Conflicts and Synergy}
\label{subsec:Asymmetry_in_Linguistic_Conflicts_and_Synergy}
We investigate linguistic conflicts and synergy during the post-training phase. To explore this, we utilize three foundation models, as mentioned in Section \ref{subsec:models}, to perform multilingual training with training datasets that include a range of languages, from 5 to 50, and evaluate the average performance on corresponding languages. 

To quantify linguistic conflicts and synergy, we compare multilingual training with separate training, where each language pair is trained independently, eliminating cross-lingual interactions.
\begin{itemize}[leftmargin=*, itemsep=0pt, parsep=0pt, topsep=0pt, partopsep=0pt]
    \item \textbf{Linguistic Conflicts}: If multilingual training underperforms compared to separate training (i.e., COMET drop), conflicts dominate over synergy.
    \item \textbf{Linguistic Synergy}: If multilingual training outperforms separate training, synergy dominates.
    \item \textbf{Intensity}: the magnitude of the performance gap measures the strength of conflicts/synergy. 
\end{itemize}

\begin{table*}[h!]
\centering
\resizebox{\linewidth}{!}{\begin{tabular}{@{}lcccccccccc@{}}
\toprule
\multirow{2}{*}{Model} & \multirow{2}{*}{Training Type} & \multicolumn{2}{c}{Group 1} & \multicolumn{2}{c}{Group 2} & \multicolumn{2}{c}{Group 3} & \multicolumn{2}{c}{Group 4} \\
\cmidrule(lr){3-4} \cmidrule(lr){5-6} \cmidrule(lr){7-8} \cmidrule(lr){9-10}
 &  & XX-En & En-XX & XX-En & En-XX & XX-En & En-XX & XX-En & En-XX \\
\midrule
\multirow{8}{*}{\makecell{ALMA-13B\\Pretrain}} & Group Training    & 86.0/30.8 & \bf 87.7/32.0 & 86.3/31.7 & \bf 87.7/33.2 & 85.2/31.5 & \bf 88.0/26.8 & 80.3/24.1 & \bf 81.3/26.0 \\
                          & Separate Training & \bf 88.5/43.4 & 86.6/29.9 & \bf 88.4/41.4 & 87.1/31.7 & \bf 86.9/39.3 & 86.9/24.7 & \bf 81.1/31.0 & 74.7/23.7 \\
                          & Multilingual      & 72.3/14.5 & 87.2/31.6 & 71.2/13.5 & 87.4/32.6 & 71.7/13.4 & 87.5/\bf 26.8 & 66.2/10.0 & 80.6/25.7 \\
\cmidrule(lr){2-10}
&& \multicolumn{2}{c}{Group 5} & \multicolumn{2}{c}{Group 6} & \multicolumn{2}{c}{Group 7} & \multicolumn{2}{c}{Group 8} \\
\cmidrule(lr){3-4} \cmidrule(lr){5-6} \cmidrule(lr){7-8} \cmidrule(lr){9-10}
 &  & XX-En & En-XX & XX-En & En-XX & XX-En & En-XX & XX-En & En-XX \\
\cmidrule(lr){2-10}
& Group Training    & 82.7/25.8 & \textbf{80.0}/16.4 & 82.7/21.9 & \textbf{ 81.1}/18.5 & 77.4/\textbf{17.7} & 68.7/9.7 & 73.6/12.6 & 69.9/6.9 \\
                          & Separate Training & \bf 83.3/29.4 &  75.6/14.2 & \bf 83.5/24.8 & 75.8/16.8 & \textbf{ 77.6}/17.6 & 57.4/5.1 & \bf 76.5/18.5 & 55.5/4.0 \\
                          & Multilingual      & 70.7/13.1 & 79.4/ \bf 16.7 & 72.3/11.1 & 80.7/ \textbf{18.7} & 62.3/4.0 & \bf 70.6/11.0 & 62.8/5.9 & \bf 70.9/8.1 \\
\midrule
&  & \multicolumn{2}{c}{Group 1} & \multicolumn{2}{c}{Group 2} & \multicolumn{2}{c}{Group 3} & \multicolumn{2}{c}{Group 4} \\
\cmidrule(lr){3-4} \cmidrule(lr){5-6} \cmidrule(lr){7-8} \cmidrule(lr){9-10}
 &  & XX-En & En-XX & XX-En & En-XX & XX-En & En-XX & XX-En & En-XX \\
 \cmidrule(lr){1-10}
\multirow{8}{*}{\makecell{X-ALMA-13B\\Pretrain}} & Group Training    & 86.3/31.1 & \bf 88.8/34.7 & 86.6/31.7 & \textbf{88.6}/33.2 & 85.9/33.2 & 89.8/31.3 & 83.5/26.6 & \bf 86.3/29.3 \\
                          & Separate Training & \bf 88.9/44.4 & 88.5/34.1 & \bf 88.6/41.6 & 88.3/34.9 & \bf 87.5/41.0 & 89.7/30.5 & \bf 85.9/35.8 & 85.3/27.8 \\
                          & Multilingual      & 79.0/17.4 & 88.6/34.3 & 78.0/16.3 & 88.5/\textbf{35.4} & 77.3/15.6 & \bf 89.9/31.4 & 75.8/13.8 & 86.2/28.8 \\
\cmidrule(lr){2-10}
&  & \multicolumn{2}{c}{Group 5} & \multicolumn{2}{c}{Group 6} & \multicolumn{2}{c}{Group 7} & \multicolumn{2}{c}{Group 8} \\
\cmidrule(lr){3-4} \cmidrule(lr){5-6} \cmidrule(lr){7-8} \cmidrule(lr){9-10}
 &  & XX-En & En-XX & XX-En & En-XX & XX-En & En-XX & XX-En & En-XX \\
 \cmidrule(lr){2-10}
 & Group Training    & 85.6/28.2 & \bf 89.9/24.0 & 86.2/24.6 & \bf 89.6/25.8 & 86.4/27.0 & \bf 81.0/18.8 & 83.0/20.6 & \bf 87.5/17.6 \\
                          & Separate Training & \bf 87.4/35.3 & 89.4/23.5 & \bf 87.7/30.1 & 88.9/22.6 & \bf 87.8/33.1 & 80.2/17.2 & \bf 86.5/31.1 & 86.6/15.6  \\
                          & Multilingual      & 79.7/17.8 & 89.7/23.6 & 80.0/14.7 & 89.0/22.8 & 77.8/10.6 & 80.9/18.3 & 76.5/11.4 & 87.2/17.1 \\
\bottomrule
\end{tabular}}
\caption{Performance of ALMA-13B-Pretrain and X-ALMA-13B-Pretrain on 50 languages from the Flores-200 test sets under three training approaches: Group multilingual training, Separate training, and Multilingual training. Results are categorized by language groups. Detailed scores for each group are provided in the Appendix.}
\label{tab:inefficiency_existing_approach}
\end{table*}

Figure \ref{fig:Asymmetric_Conflict_and_Synergy} displays the results. We can have the following observations:
\begin{itemize}[leftmargin=*, itemsep=0pt, parsep=0pt, topsep=0pt, partopsep=0pt]
    \item \textbf{Key Findings 1: Asymmetry in Linguistic Conflicts and Synergy.} As shown in Figures \ref{fig:Asymmetric_Conflict_and_Synergy} (a), (d), and (g), the average performance decreases with an increase in the number of languages, a phenomenon known as the CoM~\cite{conneau2019unsupervised,xu2024x}. However, by decomposing the average performance across all language directions into XX$\rightarrow$En and En$\rightarrow$XX, we uncover an intriguing asymmetry in the distribution of linguistic conflicts and synergies, as illustrated in Figures \ref{fig:Asymmetric_Conflict_and_Synergy} (b), (c), (e), (f), (h), and (i). Specifically, in the XX$\rightarrow$En direction, linguistic conflicts are more dominant, as shown by multilingual training consistently underperforming separate training. Conversely, in the En$\rightarrow$XX direction, linguistic synergy is significant, with multilingual tuning consistently outperforming separate training. Furthermore, comparing different models reveals that increasing model capacity (e.g., from 7B to 13B) or incorporating more languages in the pre-training corpus can mitigate conflicts. However, a significant gap remains between separate and multilingual tuning, indicating that simply increasing model capacity and the number of languages in the pre-training corpus cannot fully resolve the issue. We observe similar findings in terms of SacreBLEU (Appendix \ref{app:Asymmetry_in_Linguistic_Conflicts_and_Synergy_in_terms_of_SacreBLEU}) and across different settings (Appendix \ref{app:more_experiments_Asymmetry_in_Linguistic_Conflicts_and_Synergy}).  
    
    A potential concern regarding this phenomenon is that it may stem from the limited LoRA rank, leading to linguistic conflicts and synergy issues. However, our results (see Appendix \ref{subsec: impact_lora}) demonstrate that LoRA rank is not the root cause of this phenomenon. Instead, this issue may arise from an inherent limitation in the model’s ability to encode source language representations effectively, potentially due to the absence of an encoder component. We leave this for future work.

    \item \textbf{Key Findings 2: Multilingual Pretraining Stage can sufficiently facilitate X-ALMA-13B-Pretrain with ideal multilingual capabilities, whereas the bottleneck may lie in the post-training stage.} Observing the dotted line in Figures \ref{fig:Asymmetric_Conflict_and_Synergy} (g), (h), and (i), we find that separate training on the X-ALMA-13B-Pretrain model achieves ideal multilingual performance, maintaining average performance as the number of languages increases. However, multilingual training in the post-training stage cannot fully activate this multilingual ability, resulting in the CoM. For instance, Figure \ref{fig:Asymmetric_Conflict_and_Synergy} (h) shows a significant performance gap between multilingual training and ideal performance, which widens as the number of languages increases. Interestingly, previous work \cite{xu2024x} designed complex training regimens with up to five stages, including three pre-training and two post-training stages with language-specific group training, to address this issue. In contrast, our findings suggest there may be a more efficient way to tackle the CoM. For example, we could start with a base model that only undergoes multilingual pretraining and then apply a dedicated post-training approach to achieve high-quality translation.
\end{itemize}

\input{latex/Figure/Pipeline}

\subsection{Asymmetry in Conflicts and Synergies Across Languages Groups and Resources}
\label{subsec:Asymmetry_in_Linguistic_Conflicts_and_Synergy_Across_Language_groups_and_resources}
We further address a key question: \textit{Does Asymmetry in Conflicts and Synergies occur across all language pairs, or is it concentrated in specific pairs?} To answer this, we analyze its distribution across different language groups and resource levels.

Figure \ref{fig:performance_gap_vs_language_group_and_resource_level} displays the results. We can have the following observations:
\begin{itemize}[leftmargin=*, itemsep=0pt, parsep=0pt, topsep=0pt, partopsep=0pt]
    \item Asymmetry in linguistic conflicts is consistently observed across languages with varying resource levels and language groups, but its intensity is not uniformly distributed.
    \item While increasing model capacity or pre-training data can help narrow the performance gap, consistent with findings in previous work~\cite{arivazhagan2019massively,aharoni-etal-2019-massively,shaham-etal-2023-causes,meng-monz-2024-disentangling}, a substantial gap of nearly 10 COMET-22 points still persists.
\end{itemize}

\subsection{Challenges by Asymmetry in Linguistic Conflicts and Synergy}
\label{subsec:Challenges_by_Asymmetry_in_Linguistic_Conflicts_and_Synergy}
The asymmetry in linguistic conflicts and synergies may pose challenges for LLM-based MMT, leading to suboptimal performance for existing post-training approaches. Intuitively, translation directions where linguistic conflicts dominate may benefit from post-training strategies that minimize such conflicts. Conversely, translation directions where linguistic synergies prevail may require strategies that effectively enhance high-quality synergy. To see this, we fine-tune foundation models using three key approaches: multilingual training, group multilingual training, and separate training on 50 languages and compare their performance.

Table~\ref{tab:inefficiency_existing_approach} displays the experimental results on the Flores-200 test set. We observe the following:
\begin{itemize}[leftmargin=*, itemsep=0pt, parsep=0pt, topsep=0pt, partopsep=0pt]
    \item \textbf{Key Findings 3: The effectiveness of the existing training strategy exhibits an asymmetrical pattern.}: In XX$\rightarrow$En translations, separate training consistently achieves the best performance, followed by group multilingual training, while full multilingual training performs the worst. This result is expected, as linguistic conflict is prominent in these translation directions. 

    By contrast, in En$\rightarrow$XX translations, multilingual training or group multilingual training consistently outperforms separate training. This indicates that while linguistic conflicts dominate in the XX$\rightarrow$En direction, the En$\rightarrow$XX direction benefits from cross-linguistic knowledge transfer, leading to an enhanced translation quality. When model capacity is sufficiently large, the general pattern observed is: group multilingual training > multilingual training > separate training. This highlights two things: 1) linguistic similarity benefits positive cross-linguistic transfer. 2) the widely adopted group multilingual training approach remains insufficient to address the challenges posed by the asymmetry.
\end{itemize}

These findings underscore the critical impact of asymmetry in linguistic conflicts and synergy phenomenon on the effectiveness of existing training strategies, highlighting the need for novel training approaches to consider such an asymmetry to achieve optimal performance in both directions.

\section{Direction-Aware Training and Merging for Efficient LLM-based MMT}
\label{sec:dtam_approach}
In this section, we show how to construct an efficient MMT system by leveraging the insights from Section \ref{sec:Asymmetry_in_Linguistic_Conflicts_and_Synergy}, starting from a base model with simple multilingual pre-training.
\subsection{Motivations and Main Ideas}
As demonstrated in Section \ref{sec:Asymmetry_in_Linguistic_Conflicts_and_Synergy}, linguistic conflicts and synergy exhibit asymmetry during the post-training stage, posing significant challenges to multilingual translation. A widely adopted technique to mitigate conflicts and enhance synergy is language-specific group multilingual training~\cite{fan2021beyond,zhao2024sparse,xu-etal-2023-condensing, xu2024x}. However, it still achieves sub-optimal performance.

The state-of-the-art XALMA system~\cite{xu2024x} achieves high-quality translations by employing eight large language-specific adapters within a MoE framework combined with group multilingual training. However, this approach incurs high computational and storage costs, as each adapter contains up to 15\% of the base model’s parameters, making large-scale deployment challenging. Additionally, XALMA requires a massive amount of tokens during pre-training, further increasing resource consumption. This raises an important question: \textit{Can we achieve comparable high translation quality in a more efficient manner? }

\begin{table*}[h]
    \centering
    \resizebox{\linewidth}{!}{\begin{tabular}{@{}lccrrrcc@{}}
        \toprule
        \multirow{2}{*}{Model} & \multirow{2}{*}{\# Tokens (Pre-training)} & \multicolumn{2}{c}{\# Params} & \multicolumn{2}{c}{FLORES200} & \multicolumn{2}{c}{WMT23} \\
        \cmidrule(lr){3-4} \cmidrule(lr){5-6} \cmidrule(lr){7-8}
        & & Base/Adapter & Total  & XX$\rightarrow$En & En$\rightarrow$XX & XX$\rightarrow$En & En$\rightarrow$XX \\
        \midrule 
        \multicolumn{7}{c}{\textit{Existing State-of-the-Art MMT System }}  \\
        \midrule
        NLLB-3.3B & - & 3B/- & 3B & 80.7 & 87.4 &  71.2 &  81.5 \\
        Aya-23-8B & - & 8B/- & 8B & 80.9 & 74.4 & 81.5 &  84.2 \\
        Aya-23-35B & - & 35B/- & 35B & 84.9 & 76.0 & 82.3 &   84.1 \\
        % X-ALMA-13B (MoE) & 110B  & 13B & 8 $\times$ 1.76B  & 88.28 &  &  \bf 84.1 & \bf 85.7 \\
        Aya-101 & - & 13B/- & 13B & 86.3 & 84.1 &  79.7&  80.8 \\
        LLaMAX3-Alpaca-8B & 66B & 8B/- & 8B & 85.9 & 84.1 & 81.0 & 79.8 \\
        X-ALMA-13B (Only SFT, MoE) & 110B  & 13B/16B & 29B & \bf 88.2 & \bf 88.9 &  \bf 83.2 & \bf 85.6 \\
        \midrule
        \multicolumn{7}{c}{\textit{Our System}}  \\
        \midrule
        % X-ALMA-13B-DAT(MoE) \\ 
        X-ALMA-13B-DAT (MoE) & 20B & 13B/4B & 17B & 87.6 & 87.8 & 82.8 & 84.8 \\
        X-ALMA-13B-DATM (MoE) & 20B & 13B/1B & 14B & 87.4 & 87.8 & 82.1 & 84.8 \\
        \bottomrule
    \end{tabular}}
    \caption{Performance on Flores-200 and WMT23 benchmarks. The results of baselines are directly sourced from \citet{xu2024x} as we utilized same generation configuration. Full results are provided in Appendix.}
    \label{tab:main_results}
\end{table*}
Intuitively, we could develop a more efficient training approach for high-quality MMT by considering the asymmetry in linguistic conflicts and synergy. To this end, we propose a direction-aware training framework combined with model merging, which fully leverages the inherent asymmetry to enhance both performance and efficiency. Our approach primarily consists of two key components: 1) Direction-aware training strategies for efficiently and effectively mitigating linguistic conflicts and encouraging linguistic synergy and 2) Group-wise model merging for running efficiency. 

\subsection{Direction-Aware Training Strategies}
As shown in Figure 4 (b), we propose a simple yet effective direction-aware training strategy that addresses linguistic conflicts and linguistic synergy separately for different translation directions:
\begin{itemize}[leftmargin=*, itemsep=0pt, parsep=0pt, topsep=0pt, partopsep=0pt]
    \item For XX$\rightarrow$En translation directions: We employ separate training to build expert models for each language direction. 
    \item For En$\rightarrow$XX translation directions: We adopt group multilingual training, training one model per language group following \citet{xu2024x}.
\end{itemize}
All training employs LoRA~\cite{hu2022lora} with a rank of 16 for parameter efficiency. Using the proposed strategies, we construct a LoRA weight pool of size $N_G + N_L$, where $N_G$ is the number of groups and $N_L$ is the number of languages.

\subsection{Group-wise Model Merging}
Although the direction-aware training approach achieves promising performance, the number of LoRA weights increases linearly with the number of supported languages, posing challenges for deployment and inference, especially at large language scales. 
Model merging~\cite{yadav2024ties,zhang2023composing} provides a feasible solution to reduce the number of LoRA weights and improve efficiency. However, directly using model merge for efficient MMT is non-trial. In our preliminary experiments, we have two key observations:
\begin{itemize}[leftmargin=*, itemsep=0pt, parsep=0pt, topsep=0pt, partopsep=0pt]
    \item Merging LoRA weights into one for each direction leads to performance degradation. Notably, \citet{dang2024aya} find that model merging can improve performance, contrasting our findings. However, this discrepancy may arise because their comparison is against a weaker baseline, such as multilingual training, whereas we compare against the most vigorous baseline—separate training.
    \item \textbf{The degradation effect of model merging exhibits an asymmetric nature.} The performance degradation per parameter in the En$\rightarrow$XX direction is \textbf{6.86× greater} than in the XX$\rightarrow$En direction. A potential explanation is that linguistic synergy plays a crucial role in En$\rightarrow$XX directions, while model merging introduces low-quality linguistic synergy, leading to a performance drop.
\end{itemize}
Therefore, a more dedicated design is needed to preserve performance as much as possible.

Motivated by these observations, we only apply model merging to XX$\rightarrow$En directions in a group-wise manner. Specifically, we apply model merging to languages within each group, resulting in $N_G$ LoRA weights. We adopt the TIES~\cite{yadav2024ties} for model merging. We also compare this approach with other methods such as DARE-TIES~\cite{yu2024language} and find no significant performance difference. With this approach, we can reduce the number of LoRA weights from $\mathcal{O}(N_L)$ to $\mathcal{O}(N_G)$, improving scalability while lead minimal performance degradation.

\subsection{Main Results}
\label{subsec:Main_Results}
We evaluated our models using the Flores-200 test set for 50 languages and the WMT23 test sets for five languages (de, ru, uk, ja, zh). We provide more details in Appendix \ref{app:detailed_experimental_setups}. We select existing state-of-the-art open multilingual MT system as baselines: 
\begin{itemize}[leftmargin=*, itemsep=0pt, parsep=0pt, topsep=0pt, partopsep=0pt]
    \item \textbf{Aya-101}~\cite{ustun-etal-2024-aya}: A 13B multilingual LLM supporting 101 languages.  
    \item \textbf{LLaMAX}~\cite{lu2024llamax}: An 8B LLM-based MMT system supporting 102 languages. 
    \item \textbf{Aya-23-8B/35B}~\cite{aryabumi2024aya}: An 8B/35B multilingual LLMs that support 23 languages.
    \item \textbf{XALMA}~\cite{xu2024x}: A 29B multilingual MoE-based MMT system supporting 50 languages, using language-specific adapters and group multilingual training. Notably, since we focus only on the supervised fine-tuning stage, we select the version without preference learning, namely XALMA-13B (Only SFT) to ensure fair comparison.
\end{itemize}

Table \ref{tab:main_results} shows the results. We can have the following observations:
\begin{itemize}[leftmargin=*, itemsep=0pt, parsep=0pt, topsep=0pt, partopsep=0pt]
    \item Both X-ALMA-13B-DAT and X-ALMA-13B-DATM can achieve high translation performance. Compared to previous multilingual LLMs, such as Aya-101, Aya-23-8B, and LLaMAX, our approach consistently outperforms them across both benchmarks and translation directions. Moreover, compared to X-ALMA, our X-ALMA-13B-DAT achieves comparable performance in XX$\rightarrow$En directions; however, in En$\rightarrow$XX, a significant performance gap remains, up to 0.95 COMET-22 on average.
    \item Our approach provides an efficient way to build effective MMT. Our model is built upon X-ALMA-13B-Pretrain with only 20 billion tokens of simple multilingual pre-training. Moreover, it utilizes multiple small LoRA weight compositions and achieves relatively high translation performance across all directions, which is consistent with previous work~\cite{zheng-etal-2024-partialformer}
\end{itemize}

\section{Related Work}
\subsection{Curse of Multilinguality}
Existing research has explored both understanding and addressing this issue in MMT, identifying critical factors such as resource imbalances, limited model capacity, and complex interactions between language pairs, particularly for low-resource languages~\cite{arivazhagan2019massively,aharoni-etal-2019-massively,shaham-etal-2023-causes}. Interestingly, studies have shown that while linguistic similarity enhances positive transfer, dissimilar languages can also act as regularizers, improving training stability~\cite{meng-monz-2024-disentangling}. 
To address these challenges, proposed solutions in recent research include language-specific modules (e.g., adapters, sparse experts) to dynamically allocate capacity and reduce interference~\cite{fan2021beyond,zhao2024sparse,xu-etal-2023-condensing}, vocabulary optimization to better support new languages through improved token representations~\cite{han2024adapters}, data sampling techniques to enhance representation for underrepresented languages~\cite{wang-etal-2020-balancing,wang-neubig-2019-target,lin-etal-2019-choosing} and continual learning techniques~\cite{liu-etal-2023-continual}. Notably, techniques, such as language-specific modules, have been integrated into LLM-based MMT systems, resulting in substantial improvements in multilingual performance~\cite{xu2024x}. In this work, we systematically investigate how post-training in LLM-based MMT contributes to the CoM, providing a fine-grained analysis of its impact on linguistic conflicts and synergies.
\subsection{LLMs for Multilingual MT}
Many efforts have been made to adapt LLMs for effective machine translation. A key approach is prompting, which enhances translation performance without additional training~\cite{he-etal-2024-exploring, lu-etal-2024-chain}. Beyond this, growing research focuses on fine-tuning open and smaller LLMs to achieve high translation quality while ensuring efficiency~\cite{xu2024a, yang2023bigtranslate, alves2024tower, aryabumi2024aya}.

\citet{yang2023bigtranslate} propose a training pipeline that integrates monolingual pre-training to improve language modeling and parallel instruction fine-tuning for enhanced translation performance. Similarly, \citet{xu2024a} emphasize the quality over quantity of parallel data, introducing a training recipe: (1) large-scale monolingual pre-training, followed by (2) small-scale, high-quality parallel fine-tuning. Further revisiting the role of parallel data, \citet{guo-etal-2024-novel} highlights its importance in the pre-training stage. Additionally, \citet{xu2024contrastive} underscore the necessity of alignment in post-training, proposing the CPO algorithm. More recently, with the need to scale models across more languages, \citet{xu2024x} introduces language-specific modules combined with group training to mitigate language conflicts. In this work, we focus on the post-training stage, which has been underexplored in previous studies, and propose a direction-aware training approach with model merging to achieve efficient and effective MMT.

\section{Conclusions}
In this work, we systematically investigate linguistic conflicts and synergy during post-training in LLM-based MMT and identify a phenomenon we term asymmetry in linguistic conflicts and synergy. We provide an in-depth analysis of its distribution and challenges for LLM-based MMT. Based on these insights, we propose a direction-aware training approach combined with model merging to build an effective MMT system from X-ALMA-13B-Pretrain with only multilingual pre-training. Our approach highlights the importance of post-training in LLM-based MMT and offers insights into building MMT resource-efficiently.

\section*{Limitations}
One limitation of this work is that our approach does not surpass state-of-the-art methods like X-ALMA in performance, particularly in En$\rightarrow$XX directions, despite requiring less training cost and fewer model parameters. Second, while this work identifies a novel phenomenon and designs an efficient approach leveraging it, it does not provide a deeper or more rigorous analysis of why asymmetry in linguistic conflicts and synergy exists. We leave the analysis of the underlying mechanism of asymmetry in linguistic conflicts and synergy for future work.

Additionally, although this work conducts extensive experiments on fifty languages and three pre-trained models, further scaling is necessary to validate our findings on a broader scale, such as extending to over 100 languages. This would help push the boundaries of multilingual machine translation research, which we also leave for future work.

% \newpage
\appendix

\section{Detailed Experimental Setups}
\label{app:detailed_experimental_setups}
In this section, we will discuss the detailed setup of our experiment, including the datasets. 

\subsection{Details of Dataset in Section \ref{subsec:datasets}}

Following \cite{xu2024x}, we present a classification of languages based on linguistic families, scripts, and resource availability in Tables \ref{tab:languages1}- \ref{tab:languages2}. Fifty languages are grouped into eight distinct categories, primarily guided by linguistic similarity while considering a balanced distribution of languages across groups. Each group encompasses a mix of low-, medium-, and high-resource languages to ensure comprehensive multilingual coverage. Additionally, English is included in each group to facilitate English-centric translation and mitigate catastrophic forgetting. This structured grouping provides a well-rounded dataset for multilingual research, enabling robust language modeling and cross-lingual transfer learning.

\begin{figure*}[!ht]
    \centering
    \includegraphics[width=0.8\textwidth]{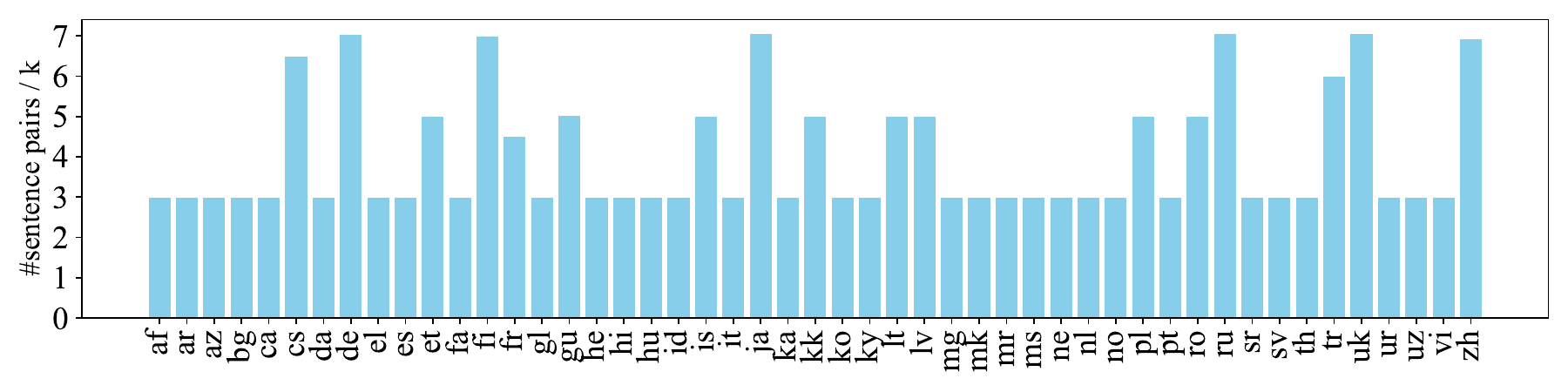}
    \vspace{-5pt}
    \caption{Number of sentences per language pair in X-ALMA-Parallel-Data \cite{xu2024x}}
    \label{fig:language_xalma_para}
\end{figure*}

We train the translation model on X-ALMA-Parallel-Data, a parallel dataset in \cite{xu2024x}.
% \footnote{\url{https://huggingface.co/datasets/haoranxu/X-ALMA-Parallel-Data}}. 
The distribution of the parallel datasets for each language is illustrated in Figure \ref{fig:language_xalma_para}.

\begin{figure*}[!ht]
    \centering
    \includegraphics[width=0.8\textwidth]{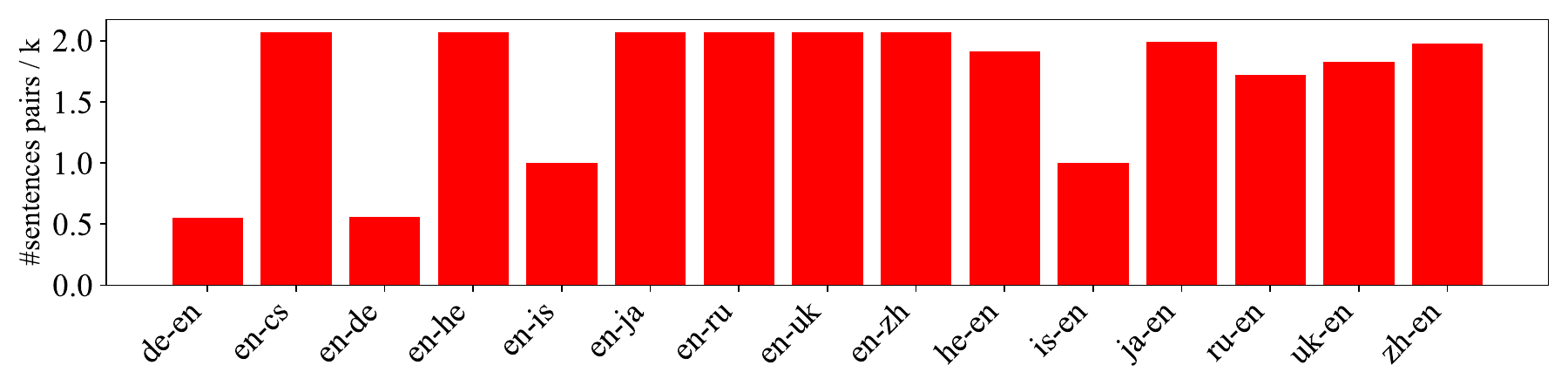}
    \vspace{-5pt}
    \caption{Number of Sentences per language pair in WMT'23}
    \label{fig:language_wmt23}
\end{figure*}

The evaluation dataset primarily consists of samples from the Flores-200 development set and NTREX~\cite{barrault2019findings}. In our experiment, we follow the setting in \cite{xu2024x}, where the translation sentences are sampled to contain 1012 sentences in each language pair.
We also use WMT23 benchmarks to assess performance for evaluation.
The distribution of WMT23 for each language is illustrated in Figure \ref{fig:language_wmt23}.

For languages in both Flores-200 and WMT’15-22, corresponding test sets are incorporated, yielding an average of 4K examples per language. 

% \subsection{Models}

\section{Additional Experiments}
\subsection{Asymmetry in Linguistic Conflicts and Synergy in terms of SacreBLEU}
\label{app:Asymmetry_in_Linguistic_Conflicts_and_Synergy_in_terms_of_SacreBLEU}

As shown in Figure \ref{fig:Asymmetric_Conflict_and_Synergy_bleu}, we observe a clear asymmetry in linguistic conflicts and synergy based on the SacreBLEU metric. This aligns with our main findings in the paper, where we used the COMET metric, further reinforcing the consistency of the observed phenomenon across different evaluation measures.

\subsection{More Experiments on Asymmetry in Linguistic Conflicts and Synergy}
\label{app:more_experiments_Asymmetry_in_Linguistic_Conflicts_and_Synergy}
We further design another setting to validate the asymmetry in linguistic conflicts and synergy.
\paragraph{Experimental Setup} We select anchor sets of varying sizes and perform post-training using training sets that include different numbers of languages but cover those anchor sets. We then observe the performance changes of these anchor sets. If the performance declines as more languages are included in the training set, this would indicate the presence of linguistic conflicts.

\paragraph{Results} Table \ref{tab:Asymmetry_results_anchor} displays the results. We can clearly observe that in the XX-En directions, the average performance of each anchor set consistently decreases as the number of languages increases. However, this phenomenon is not observed in the En-XX directions, where performance remains relatively stable. The findings are consistent with Section \ref{sec:Asymmetry_in_Linguistic_Conflicts_and_Synergy}.

\subsection{Impact of Lora Rank}
\label{subsec: impact_lora}
We observed an asymmetry in linguistic conflicts and synergies. A natural question arises: could this be due to using a low LoRA rank, which might limit learning capacity and, consequently, degrade performance? To address this concern, we selected the ALMA-13B-Pretrain model and trained it on 16 languages using different LoRA ranks, specifically 16 and 32. We then compared the performance of models with these LoRA ranks on the FLores-200 test sets. As shown in Table \ref{tab:performance_lora_complex}, increasing the LoRA rank did not yield performance improvements. Therefore, we conclude that the observed asymmetry is not attributed to using a low LoRA rank.

\section{Full Results}
% \subsection{Full Results of Section \ref{subsec:Asymmetry_in_Linguistic_Conflicts_and_Synergy}}
% \subsection{Full Results of Section \ref{subsec:Challenges_by_Asymmetry_in_Linguistic_Conflicts_and_Synergy}}
% \label{app:additional_experimental_results}
% The full experimental results of Section \ref{subsec:Challenges_by_Asymmetry_in_Linguistic_Conflicts_and_Synergy} is provided in Table \ref{tab:main_extended}, \ref{tab:main_group2_extended}, \ref{tab:main_group3_extended}, \ref{tab:main_group4_extended}, \ref{tab:main_group5_extended}, \ref{tab:main_group7_extended}, \ref{tab:main_group8_extended}. 

% \subsection{Full Results of Section \ref{subsec:Main_Results}}

\begin{table*}[ht]
    \centering
    \begin{tabular}{lccccc}
        \toprule
        \textbf{Language} & \textbf{ISO-639-1} & \textbf{Script} & \textbf{Family} & \textbf{Subgroup} & \textbf{Resource} \\
        \midrule
        English & \texttt{en} & Latin & Indo-European & Germanic & High \\
        \midrule
        \multicolumn{6}{l}{\textbf{Group 1: Germanic Languages}} \\
        Afrikaans & \texttt{af} & Latin & Indo-European & Germanic & Mid \\
        Danish & \texttt{da} & Latin & Indo-European & Germanic & Mid \\
        Dutch & \texttt{nl} & Latin & Indo-European & Germanic & High \\
        German & \texttt{de} & Latin & Indo-European & Germanic & High \\
        Icelandic & \texttt{is} & Latin & Indo-European & Germanic & Low \\
        Norwegian & \texttt{no} & Latin & Indo-European & Germanic & Low \\
        Swedish & \texttt{sv} & Latin & Indo-European & Germanic & High \\
        \midrule
        \multicolumn{6}{l}{\textbf{Group 2: Romance Languages}} \\
        Catalan & \texttt{ca} & Latin & Indo-European & Italic & High \\
        Galician & \texttt{gl} & Latin & Indo-European & Italic & Mid \\
        Italian & \texttt{it} & Latin & Indo-European & Italic & High \\
        Portuguese & \texttt{pt} & Latin & Indo-European & Italic & High \\
        Romanian & \texttt{ro} & Latin & Indo-European & Italic & Mid \\
        Spanish & \texttt{es} & Latin & Indo-European & Italic & High \\
        \midrule
        \multicolumn{6}{l}{\textbf{Group 3: Eastern and Southern Slavic Languages}} \\
        Bulgarian & \texttt{bg} & Cyrillic & Indo-European & Balto-Slavic & Mid \\
        Macedonian & \texttt{mk} & Cyrillic & Indo-European & Balto-Slavic & Low \\
        Russian & \texttt{ru} & Cyrillic & Indo-European & Balto-Slavic & High \\
        Serbian & \texttt{sr} & Cyrillic & Indo-European & Balto-Slavic & High \\
        Ukrainian & \texttt{uk} & Cyrillic & Indo-European & Balto-Slavic & Mid \\
        \midrule
        \multicolumn{6}{l}{\textbf{Group 4: Southeast Asian Languages}} \\
        French & \texttt{fr} & Latin & Indo-European & Italic & High \\
        Indonesian & \texttt{id} & Latin & Austronesian & Malayo-Polynesian & Mid \\
        Malagasy & \texttt{mg} & Latin & Austronesian & Malayo-Polynesian & Low \\
        Malay & \texttt{ms} & Latin & Austronesian & Malayo-Polynesian & Mid \\
        Thai & \texttt{th} & Thai & Tai-Kadai & Kam-Tai & Mid \\
        Vietnamese & \texttt{vi} & Latin & Austronesian & Vietic & High \\
        \bottomrule
    \end{tabular}
    \caption{Detailed information of all languages}
    \label{tab:languages1}
\end{table*}

\begin{table*}[!ht]
    \centering
\begin{tabular}{lccccc}
        \toprule
        \textbf{Language} & \textbf{ISO-639-1} & \textbf{Script} & \textbf{Family} & \textbf{Subgroup} & \textbf{Resource} \\
        \midrule
        \multicolumn{6}{l}{\textbf{Group 5: Central and Eastern European Languages}} \\
        Czech & \texttt{cs} & Latin & Indo-European & Balto-Slavic & Mid \\
        Greek & \texttt{el} & Greek & Indo-European & Graeco-Phrygian & Mid \\
        Hungarian & \texttt{hu} & Latin & Uralic & Finnic & High \\
        Latvian & \texttt{lv} & Latin & Indo-European & Balto-Slavic & Mid \\
        Lithuanian & \texttt{lt} & Latin & Indo-European & Balto-Slavic & Mid \\
        Polish & \texttt{pl} & Latin & Indo-European & Balto-Slavic & High \\
\bottomrule
    \end{tabular}
    \caption{Detailed information of all languages (cont.)}
    \label{tab:languages2}
\end{table*}

\definecolor{tiffanyblue}{RGB}{129,216,208}
\definecolor{bangdiblue}{RGB}{0,149,182}
\definecolor{kleinblue}{RGB}{0,47,167}
\definecolor{kabuliblue}{RGB}{26,85,153}
\definecolor{purple}{RGB}{138,43,226}
\usepgfplotslibrary{groupplots}
\usepgfplotslibrary{fillbetween} 
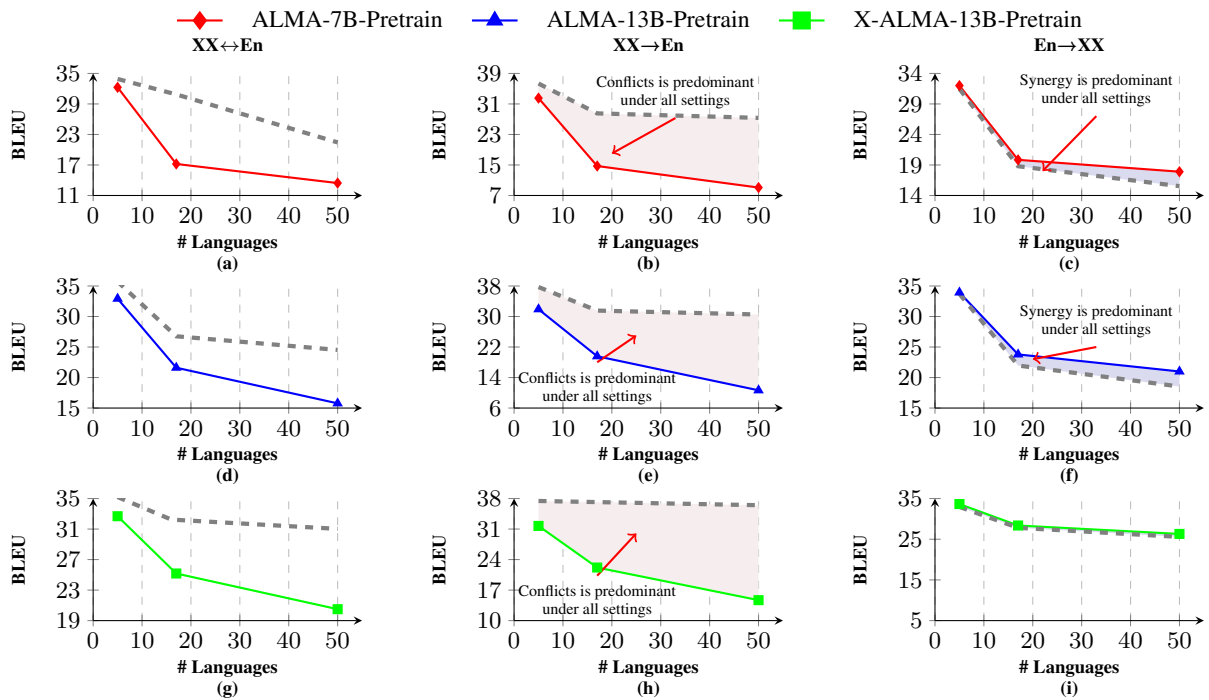
\begin{figure*}[t!]
  \centering
  \begin{tikzpicture}
    \pgfplotsset{set layers}
    \scriptsize{
    % 全局图例单独绘制
    \begin{axis}[
    at={(2,1.4)},
    width=0.5\textwidth,
      height=0.2\textwidth,
      hide axis,
      xmin=0, xmax=0.5, ymin=0, ymax=0.5,
      legend style={
        draw=none,
        fill=none,
        font=\small,
        column sep=0.3cm,
        at={(1.1,1.6)},
        anchor=north
      },
      legend columns=3
    ]
      \addlegendimage{red, mark=diamond*, mark size=3.0pt, thick, mark options={solid, fill=red}}
      \addlegendentry{ALMA-7B-Pretrain}
      \addlegendimage{blue, mark=triangle*, mark size=3.0pt, thick, mark options={solid, fill=blue}}
      \addlegendentry{ALMA-13B-Pretrain}
      \addlegendimage{green, mark=square*, mark size=3.0pt, thick, mark options={solid, fill=green}}
      \addlegendentry{X-ALMA-13B-Pretrain}
    \end{axis}
    }
    \scriptsize{
    \begin{groupplot}[
      group style={group size=3 by 3, horizontal sep=2cm, vertical sep=1.2cm},
      xmajorgrids,
      grid style={dashed, gray!50},
      width=0.32\textwidth,
      height=0.2\textwidth,
      xlabel={\# Languages},
      ylabel={BLEU},
      xlabel style={font=\bfseries, yshift=0.5em},
      ylabel style={font=\bfseries, yshift=-1.0em},
      yticklabel style={font=\small},
      xticklabel style={font=\small},
      title style={font=\bfseries},
      xmin=0, xmax=55, 
      xtick={0, 10, 20, 30, 40, 50},
        axis x line=bottom,
        axis y line=left,
        % axis y line*=left, % 左Y轴
    ]
    % 第一行：Model 1
    \nextgroupplot[title={XX$\leftrightarrow$En}, xlabel={\makecell{\# Languages \\ (a) } },ymin=11, ymax=35, ytick={11, 17, 23, 29, 35}]
    \addplot[red, mark=diamond*, mark size=1.5pt, thick, mark options={solid, fill=red}] 
      coordinates {(5, 32.262) (17, 17.199) (50, 13.442)};
    \addplot[ultra thick, dashed, gray] 
      coordinates {(5, 33.9105) (17, 30.875) (50, 21.4365)};
      
    \nextgroupplot[title={XX→En}, xlabel={\makecell{\# Languages \\ (b) } }, ymin=7, ymax=39, ytick={7, 15, 23, 31, 39}]
    \addplot[name path=mt_b,red, mark=diamond*, mark size=1.5pt, thick, mark options={solid, fill=red}] 
      coordinates {(5, 32.525) (17, 14.722) (50, 9.077)};
    \addplot[name path=separate_b,ultra thick, dashed, gray] 
      coordinates {(5, 36.348) (17, 28.524) (50, 27.344)};
    
    \addplot [
        fill=red!30!gray!10
    ] fill between [
        of=mt_b and separate_b,
    ];

    \draw[->, thick, color=red] (axis cs:33,27.3) node[above]{\tiny \textcolor{black}{\makecell{Conflicts is predominant\\ under all settings}}} -- (axis cs:20,18);

    \nextgroupplot[
        title={En→XX}, 
        xlabel={\makecell{\# Languages \\ (c) } }, 
        % 左 Y 轴（COMET-22）
        ylabel={BLEU}, 
        ymin=14, ymax=34, 
        ytick={14, 19, 24, 29, 34},
    ]
    \addplot[name path=mt_c, red, mark=diamond*, mark size=1.5pt, thick, mark options={solid, fill=red}] 
      coordinates {(5, 32.0) (17, 19.841) (50, 17.899)};
    \addplot[name path=separate_c,ultra thick, dashed, gray] 
      coordinates {(5, 31.473) (17, 18.793) (50, 15.529)};

    \addplot [
        fill=blue!40!gray!20
    ] fill between [
        of=mt_c and separate_c,
    ];

    \draw[->, thick, color=red] (axis cs:33,27) node[above]{\tiny \textcolor{black}{\makecell{Synergy is predominant\\ under all settings}}} -- (axis cs:22,18)
         ;

    % 第二行：Model 2
    \nextgroupplot[xlabel={\makecell{\# Languages \\ (d) } }, ymin=15, ymax=35, ytick={15, 20, 25, 30, 35}]
    \addplot[blue, mark=triangle*, mark size=1.5pt, thick, mark options={solid, fill=blue}] 
      coordinates {(5, 32.918) (17, 21.611) (50, 15.768)};
    
    % \addplot[ultra thick, dashed, gray] 
    %   coordinates {(5, 88.40) (9, 88.57) (17, 81.692) (50, 78.80)};

    \addplot[ultra thick, dashed, gray] 
      coordinates {(5, 35.715) (17, 26.7595) (50, 24.527)};
    
    \nextgroupplot[xlabel={\makecell{\# Languages \\ (e) } }, ymin=6, ymax=38, ytick={6, 14, 22, 30, 38}]
    \addplot[name path=mt_e, blue, mark=triangle*, mark size=1.5pt, thick, mark options={solid, fill=blue}] 
      coordinates {(5, 31.898) (17, 19.572) (50, 10.649)};

    % \addplot[ultra thick, dashed, gray] 
    %   coordinates {(5, 87.92) (9, 88.068) (17, 84.289) (50, 83.079)};
      \addplot[name path=separate_e, ultra thick, dashed, gray] 
      coordinates {(5, 37.728) (17, 31.551) (50, 30.524)};

     \addplot [
        fill=red!30!gray!10
    ] fill between [
        of=mt_e and separate_e,
    ];

    \draw[->, thick, color=red] (axis cs:17,18) node[below]{\tiny \textcolor{black}{\makecell{Conflicts is predominant\\ under all settings}}} -- (axis cs:25,25)
         ;
     
    \nextgroupplot[xlabel={\makecell{\# Languages \\ (f) } },ymin=15, ymax=35, ytick={15, 20, 25, 30, 35}]
    \addplot[name path=mt_f,blue, mark=triangle*, mark size=1.5pt, thick, mark options={solid, fill=blue}] 
      coordinates {(5, 33.938) (17, 23.787) (50, 20.993)};

     % \addplot[ultra thick, dashed, gray] 
     %  coordinates {(5, 88.882) (9, 89.071) (17, 79.095) (50, 74.52)};

      \addplot[name path=separate_f,ultra thick, dashed, gray] 
      coordinates {(5, 33.702)  (17, 21.968) (50, 18.53)};

      \addplot [
        fill=blue!40!gray!20
    ] fill between [
        of=mt_f and separate_f,
    ];

    \draw[->, thick, color=red] (axis cs:33,25) node[above]{\tiny \textcolor{black}{\makecell{Synergy is predominant\\ under all settings}}} -- (axis cs:20,23)
         ;

    % 第三行：Model 3
    \nextgroupplot[xlabel={\makecell{\# Languages \\ (g) } }, ymin=19, ymax=35, ytick={19, 23, 27, 31, 35}]
    \addplot[green, mark=square*, mark size=1.5pt, thick, mark options={solid, fill=green}] 
      coordinates {(5, 32.685) (17, 25.176) (50, 20.499)};
    \addplot[ultra thick, dashed, gray]  coordinates {(5, 35.1965) (16, 32.242) (50, 31.0265)};

    \nextgroupplot[xlabel={\makecell{\# Languages \\ (h) } }, ymin=10, ymax=38, ytick={10, 17, 24, 31, 38}]
    \addplot[name path=mt_h,green, mark=square*, mark size=1.5pt, thick, mark options={solid, fill=green}] 
      coordinates {(5, 31.72) (17, 22.175) (50, 14.693)};
    \addplot[name path=separate_h, ultra thick, dashed, gray] 
      coordinates {(5, 37.445) (50, 36.475)};

    % \node at (0.5,-10.25) {(c) XX-En};

    \addplot [
        fill=red!30!gray!10
    ] fill between [
        of=mt_h and separate_h,
    ];

    \draw[->, thick, color=red] (axis cs:17,20.3) node[below]{\tiny \textcolor{black}{\makecell{Conflicts is predominant\\ under all settings}}} -- (axis cs:25,30)
         ;
    
    \nextgroupplot[xlabel={\makecell{\# Languages \\ (i) } },ymin=5, ymax=35, ytick={5, 15,25, 35}]
    \addplot[name path=mt_i,green, mark=square*, mark size=1.5pt, thick, mark options={solid, fill=green}] 
      coordinates {(5, 33.65) (17, 28.377) (50, 26.306)};
      \addplot[name path=separate_i, ultra thick, dashed, gray]  coordinates {(5, 32.948) (17, 27.806)(50, 25.578)};

    \addplot [
        fill=blue!40!gray!20
    ] fill between [
        of=mt_i and separate_i,
    ];

    \end{groupplot}
    }
  \end{tikzpicture}
  \vskip -0.1in
  \caption{Performance of different models trained on varying numbers of languages (evaluated by SacreBLEU). The dotted line represents the performance of separately trained models, serving as a reference point where no language conflicts or synergies occur. Two key findings emerge: (1) Asymmetry in Linguistic Conflicts and Synergy (Figure a–i), highlighting the uneven impact of multilingual training across language pairs; and (2) The Bottleneck of Multilinguality in Post-Training (Figure g–i), showing that while monolingual pre-training provides an ideal foundation for handling multiple languages, the post-training stage imposes constraints that lead to the curse of multilinguality.}
  \label{fig:Asymmetric_Conflict_and_Synergy_bleu}
\end{figure*}

\begin{table*}[t!]
\centering
\resizebox{1\linewidth}{!}{%
\begin{tabular}{c c c c c}
\toprule
Languages (Trained) & Languages (Test) & XX$\rightarrow$En & En$\rightarrow$XX & All \\ 
\midrule
\multicolumn{5}{c}{\textbf{ALMA-7B-Pretrain}} \\
\midrule
\rowcolor{red!20}    5  & \{De, Zh, Ru, CS\}                         & 86.8/32.5& 88.5/32.0 & 87.7/32.3\\ 
% \rowcolor{red!20}    9  & \{De, Zh, Ru, CS\}                         & 84.97 & 88.55 & 86.76\\ 
\rowcolor{red!20}    17 & \{De, Zh, Ru, CS\}                         & 79.8/19.0 & 88.2/30.9 & 84.0/25.0\\ 
\rowcolor{red!20}    50 & \{De, Zh, Ru, CS\}                         & 76.4/17.7 & 88.4/31.3 & 82.4/24.5 \\ 
% \rowcolor{blue!20}   9  & \{De, Zh, Ru, CS, Ja, Fi, Uk, Ro\}           & 84.89 & 87.92 & 86.41\\ 
\rowcolor{blue!20}   17 & \{De, Zh, Ru, CS, Ja, Fi, Uk, Ro\}           & 79.6/18.2 & 87.4/26.5 & 83.5/22.4\\ 
\rowcolor{blue!20}   50 & \{De, Zh, Ru, CS, Ja, Fi, Uk, Ro\}           & 76.5/17.3 & 87.5/27.2 & 82.0/22.3 \\ 
\rowcolor{green!20}  17 & \{De, Zh, Ru, CS, Ja, Fi, Uk, Ro, Is, Kk, Fr, Lv, Gu, He, Hi, Hu\} & 74.3/14.7 & 78.8/19.8  & 76.6/17.3 \\
\bottomrule
\end{tabular}%
}
\caption{\textbf{The average performance of language in anchor set significantly decreases as the number of trained languages increase in XX-En directions while the average performance in En-XX directions maintain stable. This indicates the linguistic conflicts is predominant in XX-En directions, which is consistent with the findings in Section \ref{sec:Asymmetry_in_Linguistic_Conflicts_and_Synergy}.} Performance was evaluated on Flores200 test sets.}
\label{tab:Asymmetry_results_anchor}
\end{table*}

\begin{table*}[ht]
    \centering
    \begin{tabular}{l c c c c c c}
        \toprule
        \multirow{2}{*}{LoRA Rank} & \multicolumn{2}{c}{En-XX} & \multicolumn{2}{c}{XX-En} & \multicolumn{2}{c}{Avg.} \\
        \cline{2-7} 
         & COMET-22 & SacreBLEU & COMET-22 & SacreBLEU  & COMET-22 & SacreBLEU \\
        \midrule
        16 & 83.3 & 23.8 & 79.8 & 19.6 & 81.5 & 21.6\\
        32 & 83.6 & 24.1 & 79.2 & 19.6 & 81.4 & 21.8\\
        \bottomrule
    \end{tabular}
    \caption{Performance of ALMA-13B-Pretrain on FLores-200 Test Sets for Different LoRA Ranks.}
    \label{tab:performance_lora_complex}
\end{table*}

\begin{table*}[t!]
\centering
\resizebox{\linewidth}{!}{
\begin{tabular}{@{}l*{8}{c}@{\hspace{1em}}*{8}{c}@{}}
\toprule
\multicolumn{17}{c}{\textbf{Group 1 (Af, Da, Nl, De, Is, No, Sv)}} \\[0.5em]
\cmidrule(lr){1-17}
 & \multicolumn{8}{c}{XX$\rightarrow$En} & \multicolumn{8}{c}{En$\rightarrow$XX} \\
\cmidrule(lr){2-9} \cmidrule(lr){10-17}
\textbf{Strategy} & Af & Da & Nl & De & Is & No & Sv & Avg. & Af & Da & Nl & De & Is & No & Sv & Avg. \\
\midrule
\multicolumn{17}{l}{\textbf{ALMA-13B-Pretrain}} \\[0.5em]
Group Multilingual Training & 84.85 & 86.55 & 84.82 & 87.71 & 85.19 & 85.63 & 86.92 & 85.95 & 83.77 & 89.21 & 87.44 & 88.15 & 86.93 & 88.51 & 89.72 & 87.68 \\
Multilingual Training (50) & 66.14 & 70.25 & 69.87 & 79.43 & 80.38 & 69.23 & 71.04 & 72.33 & 82.65 & 88.68 & 87.01 & 87.98 & 86.80 & 88.04 & 89.22 & 87.20 \\
Separate Training & 87.56 & 89.60 & 87.38 & 89.35 & 87.04 & 88.44 & 89.79 & 88.45 & 80.94 & 88.05 & 86.81 & 87.72 & 86.89 & 87.35 & 88.58 & 86.62 \\
\midrule
\multicolumn{17}{l}{\textbf{X-ALMA-13B-Pretrain}} \\[0.5em]
Group Multilingual Training & 86.35 & 87.07 & 84.97 & 87.45 & 84.77 & 86.15 & 87.32 & 86.30 & 86.61 & 91.11 & 88.18 & 87.99 & 86.75 & 89.94 & 91.13 & 88.82 \\
Multilingual Training (50) & 76.65 & 78.71 & 76.05 & 82.53 & 82.65 & 77.29 & 78.91 & 78.97 & 86.51 & 90.93 & 88.27 & 87.77 & 86.30 & 89.81 & 90.82 & 88.63 \\
Separate Training & 89.32 & 90.23 & 87.46 & 89.28 & 86.70 & 88.93 & 90.13 & 88.86 & 86.36 & 90.53 & 87.96 & 87.66 & 86.52 & 89.65 & 90.70 & 88.48 \\
\bottomrule
\end{tabular}}
\caption{Result of various training strategies on Group 1 languages. The performance is evaluated by COMET-22.}
\label{tab:main_extended}
\end{table*}

\begin{table*}[t!]
\centering
\resizebox{\linewidth}{!}{
\begin{tabular}{@{}lccccccc@{\hspace{2em}}ccccccc@{}}
\toprule
\multicolumn{14}{c}{\textbf{Group 2 (Ca, Gl, It, Pt, Ro, Es)}} \\[0.5em]
\cmidrule(lr){1-8} \cmidrule(lr){9-15}
 & \multicolumn{7}{c}{XX$\rightarrow$En} & \multicolumn{7}{c}{En$\rightarrow$XX} \\
\cmidrule(lr){2-8} \cmidrule(lr){9-15}
\textbf{Strategy} & Ca & Gl & It & Pt & Ro & Es & Avg. & Ca & Gl & It & Pt & Ro & Es & Avg. \\
\midrule
\multicolumn{14}{l}{\textbf{ALMA-13B-Pretrain}} \\[0.5em]
Group Multilingual Training & 86.29 & 85.71 & 85.97 & 86.74 & 87.51 & 85.43 & 86.28 & 87.44 & 86.43 & 88.10 & 88.82 & 88.95 & 86.39 & 87.69 \\
Multilingual Training (50) & 69.41 & 67.79 & 70.34 & 69.30 & 82.55 & 67.91 & 71.22 & 87.14 & 85.52 & 87.94 & 88.69 & 88.67 & 86.21 & 87.36 \\
Separate Training & 88.59 & 87.88 & 88.13 & 89.33 & 89.06 & 87.42 & 88.40 & 87.10 & 84.59 & 87.83 & 88.53 & 88.61 & 86.09 & 87.12 \\
\midrule
\multicolumn{14}{l}{\textbf{X-ALMA-13B-Pretrain}} \\[0.5em]
Group Multilingual Training & 86.59 & 86.23 & 86.20 & 87.06 & 87.88 & 85.41 & 86.56 & 88.08 & 87.82 & 88.55 & 89.44 & 90.71 & 86.78 & 88.56 \\
Multilingual Training (50) & 76.67 & 76.44 & 76.87 & 77.03 & 84.77 & 75.93 & 77.95 & 88.03 & 87.72 & 88.32 & 89.36 & 90.70 & 86.68 & 88.47 \\
Separate Training & 88.81 & 88.46 & 88.18 & 89.32 & 89.36 & 87.50 & 88.61 & 87.77 & 87.59 & 88.29 & 89.21 & 90.42 & 86.51 & 88.30 \\
\bottomrule
\end{tabular}}
\caption{Result of various training strategies on Group 2 languages. The performance is evaluated by COMET-22.}
\label{tab:main_group2_extended}
\end{table*}

% \begin{table*}[t!]
% \centering
% \resizebox{\linewidth}{!}{
% \begin{tabular}{@{}lcccccc@{\hspace{2em}}cccccc@{}}
% \toprule
% \multicolumn{13}{c}{\textbf{Group 3 (Bg, Mk, Ru, Sr, Uk)}} \\[0.5em]
% \cmidrule(lr){1-7} \cmidrule(lr){8-13}
%  & \multicolumn{6}{c}{XX→En} & \multicolumn{6}{c}{En→XX} \\ 
% \cmidrule(lr){2-7} \cmidrule(lr){8-13}
% \textbf{Strategy} & Bg & Mk & Ru & Sr & Uk & Avg. & Bg & Mk & Ru & Sr & Uk & Avg. \\ 
% \midrule
% \multicolumn{13}{l}{\textbf{ALMA-13B-Pretrain}} \\[0.5em]
% Multilingual Tuning (50) & 69.35 & 66.40 & 76.99 & 68.08 & 77.54 & 71.67 & 88.47 & 83.57 & 89.44 & 86.71 & 89.16 & 87.47 \\ 
% Bilingual Tuning           & \textbf{87.45} & \textbf{85.78} & \textbf{86.86} & \textbf{87.12} & \textbf{87.26} & \textbf{86.89} & 88.21 & 81.19 & 89.57 & 86.14 & \textbf{89.21} & 86.86 \\ 
% Group Multilingual Tuning  & 85.48 & 84.30 & 85.34 & 85.44 & 85.49 & 85.21 & \textbf{89.16} & 85.24 & \textbf{89.69} & 86.80 & 89.16 & \textbf{88.01} \\ 
% \midrule
% \multicolumn{13}{l}{\textbf{X-ALMA-13B}} \\[0.5em]
% Multilingual Tuning (50) & - & - & - & - & - & - & - & - & - & - & - & - \\ 
% Bilingual Tuning           & - & - & - & - & - & - & - & - & - & - & - & - \\ 
% Group Multilingual Tuning  & - & - & - & - & - & - & - & - & - & - & - & - \\ 
% \bottomrule
% \end{tabular}
% }
% \caption{\textbf{Performance of various tuning strategies on Group 3 languages.} Results for the ALMA-13B-Pretrain model are reported, and the X-ALMA-13B results are placeholders.}
% \label{tab:main_group3_extended}
% \end{table*}

\begin{table*}[t!]
\centering
\resizebox{\linewidth}{!}{
\begin{tabular}{@{}lcccccc@{\hspace{2em}}cccccc@{}}
\toprule
\multicolumn{13}{c}{\textbf{Group 3 (Bg, Mk, Ru, Sr, Uk)}} \\[0.5em]
\cmidrule(lr){1-7} \cmidrule(lr){8-13}
 & \multicolumn{6}{c}{XX$\rightarrow$En} & \multicolumn{6}{c}{En$\rightarrow$XX} \\
\cmidrule(lr){2-7} \cmidrule(lr){8-13}
\textbf{Strategy} & Bg & Mk & Ru & Sr & Uk & Avg. & Bg & Mk & Ru & Sr & Uk & Avg. \\
\midrule
\multicolumn{13}{l}{\textbf{ALMA-13B-Pretrain}} \\[0.5em]
Group Multilingual Training & 85.48 & 84.30 & 85.34 & 85.44 & 85.49 & 85.21 & 89.16 & 85.24 & 89.69 & 86.80 & 89.16 & 88.01 \\
Multilingual Training (50) & 69.35 & 66.40 & 76.99 & 68.08 & 77.54 & 71.67 & 88.47 & 83.57 & 89.44 & 86.71 & 89.16 & 87.47 \\
Separate Training & 87.45 & 85.78 & 86.86 & 87.12 & 87.26 & 86.89 & 88.21 & 81.19 & 89.57 & 86.14 & 89.21 & 86.86 \\
\midrule
\multicolumn{13}{l}{\textbf{X-ALMA-13B-Pretrain}} \\[0.5em]
Group Multilingual Training & 86.50 & 85.77 & 85.60 & 86.00 & 85.84 & 85.94 & 90.95 & 89.55 & 89.64 & 88.74 & 90.18 & 89.81 \\
Multilingual Training (50) & 76.11 & 75.92 & 79.33 & 75.76 & 79.44 & 77.31 & 90.90 & 89.45 & 89.64 & 89.67 & 89.87 & 89.91 \\
Separate Training & 87.98 & 87.76 & 86.78 & 87.81 & 87.36 & 87.54 & 90.51 & 89.44 & 89.34 & 89.13 & 90.08 & 89.70 \\
\bottomrule
\end{tabular}}
\caption{Result of various training strategies on Group 3 languages. The performance is evaluated by COMET-22.}
\label{tab:main_group3_extended}
\end{table*}

\begin{table*}[t!]
\centering
\resizebox{\linewidth}{!}{
\begin{tabular}{@{}lccccccc@{\hspace{2em}}ccccccc@{}}
\toprule
\multicolumn{14}{c}{\textbf{Group 4 (Fr, Id, Mg, Ms, Th, Vi)}} \\[0.5em]
\cmidrule(lr){1-8} \cmidrule(lr){9-15}
 & \multicolumn{7}{c}{XX$\rightarrow$En} & \multicolumn{7}{c}{En$\rightarrow$XX} \\
\cmidrule(lr){2-8} \cmidrule(lr){9-15}
\textbf{Strategy} & Fr & Id & Mg & Ms & Th & Vi & Avg. & Fr & Id & Mg & Ms & Th & Vi & Avg. \\
\midrule
\multicolumn{14}{l}{\textbf{ALMA-13B-Pretrain}} \\[0.5em]
Group Multilingual Training & 86.98 & 85.93 & 63.47 & 84.55 & 77.34 & 83.41 & 80.28 & 87.60 & 89.92 & 63.79 & 86.85 & 72.74 & 87.01 & 81.32 \\
Multilingual Training (50) & 79.99 & 68.69 & 53.49 & 66.90 & 61.02 & 67.33 & 66.24 & 87.53 & 89.60 & 63.29 & 86.48 & 70.57 & 86.07 & 80.59 \\
Separate Training & 89.22 & 88.76 & 56.68 & 87.45 & 78.61 & 86.59 & 81.22 & 87.46 & 89.36 & 39.87 & 85.64 & 59.28 & 86.49 & 74.68 \\
\midrule
\multicolumn{14}{l}{\textbf{X-ALMA-13B-Pretrain}} \\[0.5em]
Group Multilingual Training & 86.94 & 86.35 & 73.33 & 84.82 & 85.07 & 84.24 & 83.46 & 87.96 & 90.75 & 76.37 & 87.89 & 86.51 & 88.49 & 86.33 \\
Multilingual Training (50) & 83.49 & 77.59 & 65.08 & 75.95 & 76.97 & 75.77 & 75.81 & 88.07 & 90.73 & 75.70 & 87.80 & 86.25 & 88.40 & 86.16 \\
Separate Training & 89.25 & 89.11 & 74.75 & 87.71 & 87.57 & 87.21 & 85.93 & 87.72 & 90.75 & 71.99 & 87.46 & 85.84 & 88.08 & 85.31 \\
\bottomrule
\end{tabular}}
\caption{Result of various training strategies on Group 4 languages. The performance is evaluated by COMET-22.}
\label{tab:main_group4_extended}
\end{table*}

% \begin{table*}[t!]
% \centering
% \resizebox{\linewidth}{!}{
% \begin{tabular}{@{}lccccccc@{\hspace{2em}}ccccccc@{}}
% \toprule
% \multicolumn{14}{c}{\textbf{Group 5 (Cs, El, Hu, Lv, Lt, Pl)}} \\[0.5em]
% \cmidrule(lr){1-7} \cmidrule(lr){8-14}
%  & \multicolumn{7}{c}{XX→En} & \multicolumn{7}{c}{En→XX} \\ 
% \cmidrule(lr){2-8} \cmidrule(lr){9-15}
% \textbf{Strategy} & Cs & El & Hu & Lv & Lt & Pl & Avg. & Cs & El & Hu & Lv & Lt & Pl & Avg. \\ 
% \midrule
% \multicolumn{14}{l}{\textbf{ALMA-13B-Pretrain}} \\[0.5em]
% Multilingual Tuning (50) & 79.94 & 62.88 & 68.91 & 68.85 & 63.48 & 80.26 & 70.72 & 91.11 & 75.92 & 86.24 & 68.90 & 66.40 & 87.63 & 79.37 \\ 
% Bilingual Tuning           & \textbf{88.77} & \textbf{82.72} & \textbf{87.66} & 78.75 & 75.81 & \textbf{85.94} & \textbf{83.28} & 90.95 & 68.11 & 86.23 & 61.62 & 58.89 & 87.96 & 75.63 \\ 
% Group Multilingual Tuning  & 87.92 & 81.93 & 85.57 & \textbf{79.42} & \textbf{76.79} & 84.70 & 82.72 & 91.28 & 76.51 & 86.69 & 69.81 & 67.41 & 88.21 & \textbf{79.99} \\ 
% \midrule
% \multicolumn{14}{l}{\textbf{X-ALMA-13B}} \\[0.5em]
% Multilingual Tuning (50) & - & - & - & - & - & - & - & - & - & - & - & - & - & - \\ 
% Bilingual Tuning           & - & - & - & - & - & - & - & - & - & - & - & - & - & - \\ 
% Group Multilingual Tuning  & - & - & - & - & - & - & - & - & - & - & - & - & - & - \\ 
% \bottomrule
% \end{tabular}
% }
% \caption{\textbf{Performance of various tuning strategies on Group 5 languages.} Results for the ALMA-13B-Pretrain model are reported, and the X-ALMA-13B results are placeholders.}
% \label{tab:main_group5_extended}
% \end{table*}

\begin{table*}[t!]
\centering
\resizebox{\linewidth}{!}{
\begin{tabular}{@{}lccccccc@{\hspace{2em}}ccccccc@{}}
\toprule
\multicolumn{14}{c}{\textbf{Group 5 (Cs, El, Hu, Lv, Lt, Pl)}} \\[0.5em]
\cmidrule(lr){1-8} \cmidrule(lr){9-15}
 & \multicolumn{7}{c}{XX$\rightarrow$En} & \multicolumn{7}{c}{En$\rightarrow$XX} \\
\cmidrule(lr){2-8} \cmidrule(lr){9-15}
\textbf{Strategy} & Cs & El & Hu & Lv & Lt & Pl & Avg. & Cs & El & Hu & Lv & Lt & Pl & Avg. \\
\midrule
\multicolumn{14}{l}{\textbf{ALMA-13B-Pretrain}} \\[0.5em]
Group Multilingual Training & 87.92 & 81.93 & 85.57 & 79.42 & 76.79 & 84.70 & 82.72 & 91.28 & 76.51 & 86.69 & 69.81 & 67.41 & 88.21 & 79.98 \\
Multilingual Training (50) & 79.94 & 62.88 & 68.91 & 68.85 & 63.48 & 80.26 & 70.72 & 91.11 & 75.92 & 86.24 & 68.90 & 66.40 & 87.63 & 79.37 \\
Separate Training & 88.77 & 82.72 & 87.66 & 78.75 & 75.81 & 85.94 & 83.28 & 90.95 & 68.11 & 86.23 & 61.62 & 58.89 & 87.96 & 75.63 \\
\midrule
\multicolumn{14}{l}{\textbf{X-ALMA-13B-Pretrain}} \\[0.5em]
Group Multilingual Training & 87.30 & 85.77 & 85.81 & 85.63 & 84.58 & 84.64 & 85.62 & 91.31 & 89.03 & 89.57 & 89.80 & 90.07 & 89.69 & 89.91 \\
Multilingual Training (50) & 82.44 & 76.80 & 76.78 & 81.80 & 78.77 & 81.73 & 79.72 & 91.31 & 88.88 & 89.36 & 89.49 & 89.67 & 89.70 & 89.73 \\
Separate Training & 88.58 & 87.51 & 87.99 & 87.58 & 86.49 & 86.14 & 87.38 & 90.83 & 88.87 & 89.03 & 89.64 & 89.79 & 89.61 & 89.63 \\
\bottomrule
\end{tabular}}
\caption{Result of various training strategies on Group 5 languages. The performance is evaluated by COMET-22.}
\label{tab:main_group5_extended}
\end{table*}

\begin{table*}[t!]
\centering
\resizebox{\linewidth}{!}{
\begin{tabular}{@{}lccccccc@{\hspace{2em}}cccccccc@{}}
\toprule
\multicolumn{15}{c}{\textbf{Group 6 (Et, Fi, Ja, Ka, Ko, Zh)}} \\[0.5em]
\cmidrule(lr){1-8} \cmidrule(lr){9-16}
 & \multicolumn{7}{c}{XX$\rightarrow$En} & & \multicolumn{7}{c}{En$\rightarrow$XX} \\
\cmidrule(lr){2-8} \cmidrule(lr){10-16}
\textbf{Strategy} & Et & Fi & Ja & Ka & Ko & Zh & Avg. & & Et & Fi & Ja & Ka & Ko & Zh & Avg. \\
\midrule
\multicolumn{15}{l}{\textbf{ALMA-13B-Pretrain}} \\[0.5em]
Group Multilingual Training & 80.84 & 87.60 & 86.18 & 73.71 & 85.36 & 85.22 & 83.15 &  & 73.01 & 89.85 & 89.92 & 61.63 & 84.44 & 87.75 & 81.10 \\
Multilingual Training (50) & 71.87 & 80.54 & 77.25 & 58.64 & 69.78 & 75.86 & 72.32 &  & 70.26 & 89.19 & 89.64 & 62.08 & 85.68 & 87.55 & 80.73 \\
Separate Training & 80.97 & 89.15 & 87.38 & 70.05 & 86.85 & 86.71 & 83.52 &  & 61.16 & 89.58 & 89.64 & 41.78 & 85.28 & 87.29 & 75.79 \\
\midrule
\multicolumn{15}{l}{\textbf{X-ALMA-13B-Pretrain}} \\[0.5em]
Group Multilingual Training & 87.23 & 88.25 & 86.16 & 84.34 & 85.97 & 85.01 & 86.16 &  & 90.79 & 92.29 & 90.44 & 87.02 & 88.14 & 87.46 & 89.36 \\
Multilingual Training (50) & 83.70 & 84.38 & 80.55 & 74.73 & 77.59 & 79.11 & 80.01 &  & 90.55 & 92.00 & 90.35 & 86.23 & 87.97 & 86.92 & 89.00 \\
Separate Training & 88.75 & 89.59 & 87.64 & 86.37 & 87.49 & 86.62 & 87.74 &  & 90.45 & 92.13 & 90.40 & 86.27 & 87.36 & 86.99 & 88.93 \\
\bottomrule
\end{tabular}}
\caption{Result of various training strategies on Group 6 languages. The performance is evaluated by COMET-22.}
\label{tab:main_group6_extended}
\end{table*}

% \begin{table*}[t!]
% \centering
% \resizebox{\linewidth}{!}{
% \begin{tabular}{@{}lcccccc@{\hspace{2em}}cccccc@{}}
% \toprule
% \multicolumn{13}{c}{\textbf{Group 7 (Gu, Hi, Mr, Ne, Ur)}} \\[0.5em]
% \cmidrule(lr){1-7} \cmidrule(lr){8-13}
%  & \multicolumn{6}{c}{XX→En} & \multicolumn{6}{c}{En→XX} \\ 
% \cmidrule(lr){2-7} \cmidrule(lr){8-13}
% \textbf{Strategy} & Gu & Hi & Mr & Ne & Ur & Avg. & Gu & Hi & Mr & Ne & Ur & Avg. \\ 
% \midrule
% \multicolumn{13}{l}{\textbf{ALMA-13B-Pretrain}} \\[0.5em]
% Multilingual Tuning (50) & 59.78 & 65.44 & 60.27 & 65.85 & 59.95 & 62.26 & \textbf{76.01} & \textbf{70.68} & \textbf{61.03} & \textbf{74.13} & \textbf{71.06} & \textbf{70.58} \\
% Bilingual Tuning           & 67.01 & \textbf{84.27} & \textbf{76.60} & \textbf{82.62} & \textbf{77.47} & \textbf{77.59} & 59.24 & 62.51 & 45.03 & 59.68 & 60.75 & 57.44 \\
% Group Multilingual Tuning  & \textbf{70.05} & 83.00 & 76.42 & 81.54 & 76.14 & 77.43 & 73.06 & 69.72 & 59.17 & 72.70 & 68.72 & 68.67 \\
% \midrule
% \multicolumn{13}{l}{\textbf{X-ALMA-13B}} \\[0.5em]
% Multilingual Tuning (50) & - & - & - & - & - & - & - & - & - & - & - & - \\
% Bilingual Tuning           & - & - & - & - & - & - & - & - & - & - & - & - \\
% Group Multilingual Tuning  & - & - & - & - & - & - & - & - & - & - & - & - \\
% \bottomrule
% \end{tabular}
% }
% \caption{\textbf{Performance of various tuning strategies on Group 7 languages.} Results for the ALMA-13B-Pretrain model are reported, and the X-ALMA-13B results are placeholders.}
% \label{tab:main_group7_extended}
% \end{table*}

\begin{table*}[t!]
\centering
\resizebox{\linewidth}{!}{
\begin{tabular}{@{}lcccccc@{\hspace{2em}}cccccc@{}}
\toprule
\multicolumn{13}{c}{\textbf{Group 7 (Gu, Hi, Mr, Ne, Ur)}} \\[0.5em]
\cmidrule(lr){1-7} \cmidrule(lr){8-13}
 & \multicolumn{6}{c}{XX$\rightarrow$En} & \multicolumn{6}{c}{En$\rightarrow$XX} \\
\cmidrule(lr){2-7} \cmidrule(lr){8-13}
\textbf{Strategy} & Gu & Hi & Mr & Ne & Ur & Avg. & Gu & Hi & Mr & Ne & Ur & Avg. \\
\midrule
\multicolumn{13}{l}{\textbf{ALMA-13B-Pretrain}} \\[0.5em]
Group Multilingual Training & 70.05 & 83.00 & 76.42 & 81.54 & 76.14 & 77.43 & 73.06 & 69.72 & 59.17 & 72.70 & 68.72 & 68.67 \\
Multilingual Training (50) & 59.78 & 65.44 & 60.27 & 65.85 & 59.95 & 62.26 & 76.01 & 70.68 & 61.03 & 74.13 & 71.06 & 70.58 \\
Separate Training & 66.96 & 84.27 & 76.60 & 82.62 & 77.47 & 77.59 & 59.24 & 62.51 & 45.03 & 59.68 & 60.75 & 57.44 \\
\midrule
\multicolumn{13}{l}{\textbf{X-ALMA-13B-Pretrain}} \\[0.5em]
Group Multilingual Training & 86.43 & 87.19 & 85.57 & 87.82 & 84.84 & 86.37 & 86.68 & 79.15 & 74.19 & 82.71 & 82.09 & 80.96 \\
Multilingual Training (50) & 79.75 & 77.66 & 76.45 & 79.23 & 75.84 & 77.79 & 86.60 & 79.37 & 73.76 & 82.29 & 82.21 & 80.85 \\
Separate Training & 87.07 & 88.83 & 87.44 & 89.29 & 86.49 & 87.82 & 85.91 & 78.57 & 73.17 & 81.71 & 81.84 & 80.24 \\
\bottomrule
\end{tabular}}
\caption{Result of various training strategies on Group 7 languages. The performance is evaluated by COMET-22.}
\label{tab:main_group7_extended}
\end{table*}

\begin{table*}[t!]
\centering
\resizebox{\linewidth}{!}{
\begin{tabular}{@{}lccccccccc@{\hspace{2em}}lcccccccccc@{}}
\toprule
\multicolumn{20}{c}{\textbf{Group 8 (Ar, Az, He, Kk, Ky, Fa, Tr, Uz)}} \\[0.5em]
\cmidrule(lr){1-10} \cmidrule(lr){11-21}
 & \multicolumn{9}{c}{XX$\rightarrow$En} & & \multicolumn{9}{c}{En$\rightarrow$XX} \\
\cmidrule(lr){2-10} \cmidrule(lr){12-21}
\textbf{Strategy} & Ar & Az & He & Kk & Ky & Fa & Tr & Uz & Avg. & & & Ar & Az & He & Kk & Ky & Fa & Tr & Uz & Avg. \\
\midrule
\multicolumn{20}{l}{\textbf{ALMA-13B-Pretrain}} \\[0.5em]
Group Multilingual Training & 74.17 & 72.83 & 70.54 & 73.58 & 70.06 & 75.22 & 82.01 & 70.34 & 73.59 &  &  & 77.14 & 63.41 & 71.63 & 68.57 & 62.74 & 71.52 & 77.61 & 66.83 & 69.93 \\
Multilingual Training (50) & 60.33 & 61.77 & 57.48 & 63.74 & 60.45 & 62.40 & 76.40 & 59.61 & 62.77 &  &  & 75.78 & 65.72 & 71.12 & 71.08 & 63.12 & 72.18 & 78.45 & 69.64 & 70.89 \\
Separate Training & 80.97 & 75.94 & 77.71 & 72.45 & 67.25 & 81.80 & 85.36 & 70.53 & 76.50 &  &  & 73.00 & 43.13 & 62.30 & 46.70 & 35.71 & 64.67 & 74.37 & 44.37 & 55.53 \\
\midrule
\multicolumn{20}{l}{\textbf{X-ALMA-13B-Pretrain}} \\[0.5em]
Group Multilingual Training & 81.70 & 82.67 & 83.28 & 83.53 & 81.19 & 83.14 & 87.15 & 81.15 & 82.98 &  &  & 86.40 & 86.86 & 87.74 & 89.41 & 87.42 & 86.69 & 88.50 & 86.80 & 87.48 \\
Multilingual Training (50) & 74.48 & 75.55 & 75.45 & 78.74 & 74.45 & 75.96 & 83.65 & 73.35 & 76.45 &  &  & 86.11 & 86.67 & 87.44 & 89.33 & 87.22 & 86.60 & 88.35 & 86.20 & 87.24 \\
Separate Training & 86.57 & 85.65 & 87.77 & 87.00 & 84.49 & 87.23 & 88.83 & 84.21 & 86.47 &  &  & 85.72 & 85.81 & 87.88 & 88.84 & 86.45 & 86.44 & 87.94 & 83.71 & 86.60 \\
\bottomrule
\end{tabular}}
\caption{Result of various training strategies on Group 8 languages. The performance is evaluated by COMET-22.}
\label{tab:main_group8_extended}
\end{table*}

\begin{table*}[t!]
\centering
\resizebox{\linewidth}{!}{
\begin{tabular}{@{}l cccccccc@{\hspace{2em}}ccccccccc@{}}
\toprule
\multicolumn{17}{c}{\textbf{Group 1 (Af, Da, Nl, De, Is, No, Sv)}} \\[0.5em]
\cmidrule(lr){1-9} \cmidrule(lr){10-18}
 & \multicolumn{8}{c}{XX$\rightarrow$En} & & \multicolumn{8}{c}{En$\rightarrow$XX} \\
\cmidrule(lr){2-9} \cmidrule(lr){11-18}
\textbf{Strategy} & Af & Da & Nl & De & Is & No & Sv & Avg. & & Af & Da & Nl & De & Is & No & Sv & Avg. \\
\midrule
\multicolumn{17}{l}{\textbf{ALMA-13B-Pretrain}} \\[0.5em]
Group Multilingual Training & 35.39 & 31.82 & 23.66 & 34.88 & 28.67 & 29.25 & 31.76 & 30.78 &  & 33.13 & 38.35 & 24.66 & 36.48 & 25.37 & 27.93 & 37.83 & 31.96 \\
Multilingual Training (50) & 10.68 & 12.10 & 10.23 & 21.00 & 24.37 & 11.10 & 12.09 & 14.51 &  & 33.00 & 36.93 & 24.34 & 36.82 & 25.07 & 27.65 & 37.18 & 31.57 \\
Separate Training & 52.88 & 46.74 & 32.21 & 44.21 & 37.39 & 42.92 & 47.21 & 43.37 &  & 27.63 & 35.25 & 23.39 & 36.22 & 25.41 & 25.40 & 35.72 & 29.86 \\
\midrule
\multicolumn{17}{l}{\textbf{X-ALMA-13B-Pretrain}} \\[0.5em]
Group Multilingual Training & 37.40 & 32.39 & 23.63 & 33.66 & 28.30 & 29.68 & 32.74 & 31.11 &  & 40.68 & 42.78 & 26.38 & 36.09 & 24.23 & 30.85 & 41.80 & 34.69 \\
Multilingual Training (50) & 16.45 & 16.06 & 11.58 & 22.41 & 24.56 & 14.76 & 16.14 & 17.42 &  & 40.21 & 41.95 & 26.52 & 35.32 & 23.82 & 30.57 & 41.79 & 34.31 \\
Separate Training & 57.35 & 48.92 & 32.66 & 43.83 & 35.92 & 43.87 & 48.54 & 44.44 &  & 39.08 & 42.41 & 26.08 & 35.04 & 24.05 & 30.50 & 41.25 & 34.06 \\
\bottomrule
\end{tabular}}
\caption{Result of various training strategies on Group 1 languages. The performance is evaluated by SacreBLEU.}
\label{tab:main_group1_extended}
\end{table*}

\begin{table*}[t!]
\centering
\resizebox{\linewidth}{!}{
\begin{tabular}{@{}l ccccccc@{\hspace{2em}}ccccccccc@{}}
\toprule
\multicolumn{15}{c}{\textbf{Group 2 (Ca, Gl, It, Pt, Ro, Es)}} \\[0.5em]
\cmidrule(lr){1-8} \cmidrule(lr){9-16}
 & \multicolumn{7}{c}{XX$\rightarrow$En} & & \multicolumn{7}{c}{En$\rightarrow$XX} \\
\cmidrule(lr){2-8} \cmidrule(lr){10-16}
\textbf{Strategy} & Ca & Gl & It & Pt & Ro & Es & Avg. & & Ca & Gl & It & Pt & Ro & Es & Avg. \\
\midrule
\multicolumn{15}{l}{\textbf{ALMA-13B-Pretrain}} \\[0.5em]
Group Multilingual Training & 33.77 & 31.56 & 27.82 & 34.82 & 35.49 & 26.81 & 31.71 &  & 39.10 & 31.01 & 27.53 & 41.43 & 32.21 & 27.81 & 33.18 \\
Multilingual Training (50) & 11.95 & 10.61 & 11.07 & 11.14 & 27.33 & 9.09 & 13.53 &  & 38.50 & 29.64 & 27.24 & 40.89 & 32.24 & 27.02 & 32.59 \\
Separate Training & 46.06 & 41.02 & 35.75 & 47.77 & 43.82 & 33.89 & 41.38 &  & 37.44 & 27.43 & 26.37 & 40.84 & 31.85 & 26.38 & 31.72 \\
\midrule
\multicolumn{15}{l}{\textbf{X-ALMA-13B-Pretrain}} \\[0.5em]
Group Multilingual Training & 33.79 & 31.43 & 27.45 & 34.99 & 35.41 & 26.29 & 31.56 &  & 41.11 & 35.11 & 29.31 & 43.43 & 37.63 & 28.15 & 35.79 \\
Multilingual Training (50) & 14.74 & 13.75 & 13.19 & 14.84 & 28.91 & 12.40 & 16.30 &  & 40.59 & 34.85 & 29.10 & 43.23 & 36.83 & 28.02 & 35.44 \\
Separate Training & 46.01 & 42.02 & 35.54 & 47.73 & 44.62 & 33.79 & 41.62 &  & 39.45 & 33.71 & 28.55 & 43.09 & 36.86 & 27.55 & 34.87 \\
\bottomrule
\end{tabular}}
\caption{Result of various training strategies on Group 2 languages. The performance is evaluated by SacreBLEU.}
\label{tab:main_group2_extended}
\end{table*}

\begin{table*}[t!]
\centering
\resizebox{\linewidth}{!}{
\begin{tabular}{@{}l cccccc@{\hspace{2em}}cccccccc@{}}
\toprule
\multicolumn{13}{c}{\textbf{Group 3 (Bg, Mk, Ru, Sr, Uk)}} \\[0.5em]
\cmidrule(lr){1-7} \cmidrule(lr){8-14}
 & \multicolumn{6}{c}{XX$\rightarrow$En} & & \multicolumn{6}{c}{En$\rightarrow$XX} \\
\cmidrule(lr){2-7} \cmidrule(lr){9-14}
\textbf{Strategy} & Bg & Mk & Ru & Sr & Uk & Avg. &  & Bg & Mk & Ru & Sr & Uk & Avg. \\
\midrule
\multicolumn{13}{l}{\textbf{ALMA-13B-Pretrain}} \\[0.5em]
Group Multilingual Training & 30.72 & 31.47 & 30.09 & 33.53 & 31.89 & 31.54 &  & 30.05 & 22.85 & 29.03 & 26.63 & 25.21 & 26.75 \\
Multilingual Training (50) & 9.96 & 9.16 & 18.08 & 10.21 & 19.70 & 13.42 &  & 30.21 & 21.87 & 29.62 & 26.79 & 25.28 & 26.75 \\
Separate Training & 39.56 & 38.58 & 36.13 & 42.05 & 40.32 & 39.33 &  & 28.31 & 17.45 & 29.18 & 23.40 & 25.11 & 24.69 \\
\midrule
\multicolumn{13}{l}{\textbf{X-ALMA-13B-Pretrain}} \\[0.5em]
Group Multilingual Training & 33.56 & 34.13 & 30.92 & 34.43 & 32.81 & 33.17 &  & 36.66 & 32.41 & 28.98 & 31.32 & 27.09 & 31.29 \\
Multilingual Training (50) & 12.96 & 13.49 & 18.33 & 14.14 & 18.85 & 15.55 &  & 36.36 & 32.30 & 29.38 & 32.00 & 26.90 & 31.39 \\
Separate Training & 40.40 & 43.17 & 36.69 & 44.00 & 40.71 & 40.99 &  & 35.61 & 31.28 & 28.61 & 30.38 & 26.45 & 30.47 \\
\bottomrule
\end{tabular}}
\caption{Result of various training strategies on Group 3 languages. The performance is evaluated by SacreBLEU.}
\label{tab:main_group3_extended}
\end{table*}
~

\begin{table*}[t!]
\centering
\resizebox{\linewidth}{!}{
\begin{tabular}{@{}l ccccccc@{\hspace{2em}}ccccccccc@{}}
\toprule
\multicolumn{15}{c}{\textbf{Group 4 (Fr, Id, Mg, Ms, Th, Vi)}} \\[0.5em]
\cmidrule(lr){1-8} \cmidrule(lr){9-16}
 & \multicolumn{7}{c}{XX$\rightarrow$En} & & \multicolumn{7}{c}{En$\rightarrow$XX} \\
\cmidrule(lr){2-8} \cmidrule(lr){10-16}
\textbf{Strategy} & Fr & Id & Mg & Ms & Th & Vi & Avg. & & Fr & Id & Mg & Ms & Th & Vi & Avg. \\
\midrule
\multicolumn{15}{l}{\textbf{ALMA-13B-Pretrain}} \\[0.5em]
Group Multilingual Training & 35.54 & 30.22 & 10.13 & 29.61 & 13.75 & 25.54 & 24.13 &  & 43.95 & 37.60 & 4.01 & 28.67 & 6.73 & 34.86 & 25.97 \\
Multilingual Training (50) & 24.53 & 9.34 & 3.85 & 9.18 & 4.22 & 8.58 & 9.95 &  & 44.11 & 36.40 & 4.38 & 28.37 & 7.15 & 33.54 & 25.66 \\
Separate Training & 44.78 & 41.77 & 7.63 & 40.57 & 16.36 & 34.66 & 30.96 &  & 43.97 & 35.24 & 54.00 & 25.13 & 4.11 & 32.97 & 23.66 \\
\midrule
\multicolumn{15}{l}{\textbf{X-ALMA-13B-Pretrain}} \\[0.5em]
Group Multilingual Training & 35.04 & 31.29 & 16.10 & 29.44 & 21.36 & 26.20 & 26.57 &  & 46.13 & 39.97 & 9.99 & 30.83 & 9.93 & 38.68 & 29.25 \\
Multilingual Training (50) & 27.13 & 13.59 & 6.83 & 13.09 & 10.01 & 12.26 & 13.82 &  & 45.36 & 40.01 & 9.23 & 29.28 & 11.36 & 37.54 & 28.80 \\
Separate Training & 44.69 & 42.99 & 19.64 & 40.79 & 29.92 & 36.98 & 35.84 &  & 45.30 & 39.49 & 6.48 & 28.01 & 9.78 & 37.43 & 27.75 \\
\bottomrule
\end{tabular}}
\caption{Result of various training strategies on Group 4 languages. The performance is evaluated by SacreBLEU.}
\label{tab:main_group4_extended}
\end{table*}~

\begin{table*}[t!]
\centering
\resizebox{\linewidth}{!}{
\begin{tabular}{@{}l ccccccc@{\hspace{2em}}cccccccc@{}}
\toprule
\multicolumn{15}{c}{\textbf{Group 5 (Cs, El, Hu, Lv, Lt, Pl)}} \\[0.5em]
\cmidrule(lr){1-8} \cmidrule(lr){9-16}
 & \multicolumn{7}{c}{XX$\rightarrow$En} & & \multicolumn{7}{c}{En$\rightarrow$XX} \\
\cmidrule(lr){2-8} \cmidrule(lr){10-16}
\textbf{Strategy} & Cs & El & Hu & Lv & Lt & Pl & Avg. &  & Cs & El & Hu & Lv & Lt & Pl & Avg. \\
\midrule
\multicolumn{15}{l}{\textbf{ALMA-13B-Pretrain}} \\[0.5em]
Group Multilingual Training & 36.91 & 23.58 & 26.31 & 21.62 & 19.37 & 27.05 & 25.81 &  & 31.04 & 13.04 & 17.21 & 10.46 & 8.63 & 18.20 & 16.43 \\
Multilingual Training (50) & 22.08 & 6.40 & 8.19 & 11.58 & 8.37 & 21.72 & 13.06 &  & 31.02 & 13.59 & 17.70 & 10.33 & 9.68 & 18.14 & 16.74 \\
Separate Training & 41.62 & 27.38 & 33.58 & 22.89 & 19.97 & 30.96 & 29.40 &  & 30.62 & 9.04 & 15.56 & 6.55 & 5.59 & 17.93 & 14.22 \\
\midrule
\multicolumn{15}{l}{\textbf{X-ALMA-13B-Pretrain}} \\[0.5em]
Group Multilingual Training & 34.26 & 28.15 & 25.25 & 28.89 & 26.61 & 26.19 & 28.22 &  & 30.91 & 23.62 & 21.91 & 24.56 & 22.52 & 20.28 & 23.97 \\
Multilingual Training (50) & 23.16 & 12.64 & 11.68 & 21.52 & 16.20 & 21.59 & 17.80 &  & 30.23 & 23.40 & 21.72 & 23.96 & 22.27 & 20.02 & 23.60 \\
Separate Training & 40.95 & 35.56 & 34.62 & 35.99 & 33.79 & 31.12 & 35.34 &  & 29.92 & 22.18 & 21.17 & 25.27 & 22.27 & 20.05 & 23.48 \\
\bottomrule
\end{tabular}}
\caption{Result of various training strategies on Group 5 languages. The performance is evaluated by SacreBLEU.}
\label{tab:main_group5_extended}
\end{table*}

\begin{table*}[t!]
\centering
\resizebox{\linewidth}{!}{
\begin{tabular}{@{}lccccccc@{\hspace{2em}}cccccccc@{}}
\toprule
\multicolumn{15}{c}{\textbf{Group 6 (Et, Fi, Ja, Ka, Ko, Zh)}} \\[0.5em]
\cmidrule(lr){1-8} \cmidrule(lr){9-16}
 & \multicolumn{7}{c}{XX$\rightarrow$En} & & \multicolumn{7}{c}{En$\rightarrow$XX} \\
\cmidrule(lr){2-8} \cmidrule(lr){10-16}
\textbf{Strategy} & Et & Fi & Ja & Ka & Ko & Zh & Avg. & & Et & Fi & Ja & Ka & Ko & Zh & Avg. \\
\midrule
\multicolumn{15}{l}{\textbf{ALMA-13B-Pretrain}} \\[0.5em]
Group Multilingual Training & 22.49 & 27.83 & 22.64 & 12.35 & 21.50 & 23.92 & 21.79 &  & 10.60 & 18.59 & 29.63 & 4.40 & 6.79 & 40.49 & 18.42 \\
Multilingual Training (50) & 13.25 & 18.25 & 11.69 & 3.56 & 7.05 & 12.82 & 11.10 &  & 10.10 & 18.29 & 28.71 & 5.34 & 9.87 & 39.74 & 18.67 \\
Separate Training & 24.10 & 33.33 & 26.70 & 9.22 & 26.77 & 28.95 & 24.84 &  & 5.44 & 17.45 & 28.49 & 1.52 & 9.06 & 38.79 & 16.79 \\
\midrule
\multicolumn{15}{l}{\textbf{X-ALMA-13B-Pretrain}} \\[0.5em]
Group Multilingual Training & 29.46 & 27.76 & 22.32 & 20.23 & 22.24 & 22.97 & 24.16  &  & 21.92 & 22.78 & 31.87 & 12.59 & 10.08 & 39.99 & 23.21 \\
Multilingual Training (50) & 22.32 & 20.01 & 12.44 & 9.06 & 10.38 & 14.09 & 14.72 &  & 21.48 & 21.38 & 30.89 & 12.32 & 11.82 & 38.65 & 22.76 \\
Separate Training & 35.84 & 34.19 & 26.94 & 27.00 & 28.37 & 28.31 & 30.11 &  & 21.52 & 21.62 & 31.76 & 11.54 & 10.65 & 38.22 & 22.55 \\
\bottomrule
\end{tabular}}
\caption{Result of various training strategies on Group 6 languages. The performance is evaluated by SacreBLEU.}
\label{tab:main_group6_extended}
\end{table*}

\begin{table*}[t!]
\centering
\resizebox{\linewidth}{!}{
\begin{tabular}{@{}l cccccc@{\hspace{2em}}ccccccc@{}}
\toprule
\multicolumn{13}{c}{\textbf{Group 7 (Gu, Hi, Mr, Ne, Ur)}} \\[0.5em]
\cmidrule(lr){1-7} \cmidrule(lr){8-14}
 & \multicolumn{6}{c}{XX$\rightarrow$En} & & \multicolumn{6}{c}{En$\rightarrow$XX} \\
\cmidrule(lr){2-7} \cmidrule(lr){9-14}
\textbf{Strategy} & Gu & Hi & Mr & Ne & Ur & Avg. & & Gu & Hi & Mr & Ne & Ur & Avg. \\
\midrule
\multicolumn{13}{l}{\textbf{ALMA-13B-Pretrain}} \\[0.5em]
Group Multilingual Training & 10.95 & 24.45 & 16.88 & 19.50 & 16.89 & 17.73 &  & 8.78 & 15.68 & 5.63 & 8.67 & 9.90 & 9.73 \\
Multilingual Training (50) & 3.10 & 5.34 & 3.71 & 4.46 & 3.59 & 4.04 &  & 10.50 & 17.29 & 6.25 & 9.53 & 11.33 & 10.98 \\
Separate Training & 8.09 & 26.64 & 15.40 & 20.37 & 17.67 & 17.63 &  & 4.89 & 9.66 & 2.09 & 3.89 & 5.11 & 5.13 \\
\midrule
\multicolumn{13}{l}{\textbf{X-ALMA-13B-Pretrain}} \\[0.5em]
Group Multilingual Training & 26.47 & 29.68 & 25.56 & 27.99 & 25.45 & 27.03 &  & 18.18 & 26.33 & 13.57 & 16.75 & 18.94 & 18.75 \\
Multilingual Training (50) & 12.55 & 10.32 & 10.44 & 10.05 & 9.80 & 10.63 &  & 17.78 & 25.96 & 13.17 & 15.90 & 18.88 & 18.34 \\
Separate Training & 30.82 & 36.37 & 32.69 & 33.91 & 31.51 & 33.06 &  & 16.55 & 23.83 & 12.58 & 15.01 & 17.78 & 17.15 \\
\bottomrule
\end{tabular}}
\caption{Result of various training strategies on Group 7 languages. The performance is evaluated by SacreBLEU.}
\label{tab:main_group7_extended}
\end{table*}

\begin{table*}[t!]
\centering
\resizebox{\linewidth}{!}{
\begin{tabular}{@{}l ccccccccc@{\hspace{2em}}ccccccccccc@{}}
\toprule
\multicolumn{19}{c}{\textbf{Group 8 (Ar, Az, He, Kk, Ky, Fa, Tr, Uz)}} \\[0.5em]
\cmidrule(lr){1-10} \cmidrule(lr){11-20}
 & \multicolumn{9}{c}{XX$\rightarrow$En} & & \multicolumn{9}{c}{En$\rightarrow$XX} \\
\cmidrule(lr){2-10} \cmidrule(lr){12-20}
\textbf{Strategy} & Ar & Az & He & Kk & Ky & Fa & Tr & Uz & Avg. & & Ar & Az & He & Kk & Ky & Fa & Tr & Uz & Avg. \\
\midrule
\multicolumn{19}{l}{\textbf{ALMA-13B-Pretrain}} \\[0.5em]
Group Multilingual Training & 14.67 & 9.20 & 14.01 & 10.83 & 7.40 & 14.42 & 21.09 & 9.47 & 12.64 &  & 10.04 & 3.34 & 10.70 & 4.42 & 2.41 & 11.27 & 9.99 & 3.16 & 6.92 \\
Multilingual Training (50) & 5.09 & 3.87 & 4.01 & 5.63 & 3.46 & 5.18 & 15.65 & 4.19 & 5.88 &  & 11.04 & 4.76 & 10.80 & 5.87 & 2.87 & 12.39 & 11.96 & 5.05 & 8.09 \\
Separate Training & 28.08 & 12.64 & 25.00 & 11.36 & 6.59 & 24.96 & 28.39 & 10.72 & 18.47 &  & 8.03 & 84.00 & 6.60 & 1.13 & 40.00 & 6.09 & 7.95 & 53.00 & 3.95 \\
\midrule
\multicolumn{19}{l}{\textbf{X-ALMA-13B-Pretrain}} \\[0.5em]
Group Multilingual Training & 21.40 & 16.27 & 24.74 & 19.94 & 14.91 & 20.98 & 28.69 & 17.76 & 20.59 &  & 20.13 & 11.89 & 25.11 & 17.26 & 11.53 & 21.95 & 21.56 & 10.98 & 17.55 \\
Multilingual Training (50) & 11.13 & 7.89 & 11.92 & 12.76 & 8.08 & 10.83 & 20.84 & 7.77 & 11.40 &  & 19.43 & 12.11 & 24.68 & 17.22 & 10.65 & 21.66 & 20.39 & 10.52 & 17.08 \\
Separate Training & 38.34 & 22.16 & 41.85 & 30.35 & 20.97 & 33.66 & 35.90 & 25.28 & 31.06 &  & 18.84 & 10.10 & 24.62 & 15.62 & 8.89 & 20.92 & 19.41 & 6.56 & 15.62 \\
\bottomrule
\end{tabular}}
\caption{Result of various training strategies on Group 8 languages. The performance is evaluated by SacreBLEU.}
\label{tab:main_group8_extended}
\end{table*}

\begin{table*}[h]
    \centering
    \resizebox{\linewidth}{!}{%
    \begin{tabular}{@{}lcccccccccccc@{}}
        \toprule
        \multirow{2}{*}{Model} & \multicolumn{2}{c}{Zh} & \multicolumn{2}{c}{De} & \multicolumn{2}{c}{Ru} & \multicolumn{2}{c}{Ja} & \multicolumn{2}{c}{Uk} & \multicolumn{2}{c}{Avg.} \\
        \cmidrule(lr){2-3} \cmidrule(lr){4-5} \cmidrule(lr){6-7} \cmidrule(lr){8-9} \cmidrule(lr){10-11} \cmidrule(lr){12-13}
        & BLEU & COMET & BLEU & COMET & BLEU & COMET & BLEU & COMET & BLEU & COMET & BLEU & COMET \\
        \midrule
        NLLB-3.3B &  11.4 & 67.8  &  20.1 & 66.6 & 24.4 & 76.7 &  6.8 & 65.8  &  33.1  & 79.0  &19.2  & 71.2 \\
        Aya-23-8B & 22.6 & 78.8  &  32.3 & 82.1  & 30.9 & 81.6 &  19.8  & 80.2 &  39.2  & 85.0 & 29.0  & 81.5 \\
        Aya-23-35B & 23.5 & 79.7 & 32.7 & 82.3 & 31.7 & 82.2 & 21.3 & 81.6 &  39.1 & 85.7 & 29.7 & 82.3 \\
        Aya-101 &  13.8 & 73.7 &  34.9 & 81.6 & 28.4 & 81.4 &  13.9 & 77.3 & 34.9 & 84.5 &  25.2 & 79.7  \\
        LLaMAX3-Alpaca-8B & 22.3 & 79.3 &  25.6 & 79.4 & 29.4 & 81.3 &  17.6 & 80.1 &  37.8 & 84.9 & 26.5 & 81.0\\
        X-ALMA-13B (Only SFT, MoE) & 23.8 & 80.3 & 42.5 & 85.3  &  32.8 & 82.4  &  20.4  &  81.6&  42.5  & 86.4 &  32.4 & 83.2 \\
        \midrule
        \textbf{X-ALMA-13B-DAT (MoE)} & 21.1 & 79.7 & 38.4 & 84.6 & 32.3 & 82.4 & 19.8 & 81.2 & 39.5 & 85.9 & 30.2 & 82.8 \\
        \textbf{X-ALMA-13B-DATM (MoE)} & 22.2  & 79.2 & 37.5  &  84.2 & 29.0 & 82.0  & 19.0 & 80.4 & 34.0 & 84.7 & 28.3 & 82.1 \\
        \bottomrule
    \end{tabular}}
    \caption{WMT23 XX$\rightarrow$En translation results (BLEU and COMET). The results of baselines are directly sourced from \citet{xu2024x}. We keep the same generation configuration as \citet{xu2024x}.}
    \label{tab:wmt_xx_en}
\end{table*}

\begin{table*}[h]
    \centering
    \resizebox{\linewidth}{!}{%
    \begin{tabular}{@{}lcccccccccccc@{}}
        \toprule
        \multirow{2}{*}{Model} & \multicolumn{2}{c}{Zh} & \multicolumn{2}{c}{De} & \multicolumn{2}{c}{Ru} & \multicolumn{2}{c}{Ja} & \multicolumn{2}{c}{Uk} & \multicolumn{2}{c}{Avg.} \\
        \cmidrule(lr){2-3} \cmidrule(lr){4-5} \cmidrule(lr){6-7} \cmidrule(lr){8-9} \cmidrule(lr){10-11} \cmidrule(lr){12-13}
        & BLEU & COMET & BLEU & COMET & BLEU & COMET & BLEU & COMET & BLEU & COMET & BLEU & COMET \\
        \midrule
        NLLB-3.3B &  34.8 & 79.6  &  33.5 &79.7& 29.1 & 83.8 &   13.8 & 81.6  &   25.5 &  82.8  &27.3  & 81.5 \\
        Aya-23-8B &  44.5 & 85.3  &  29.3 & 80.4  & 24.3 & 84.3 &   19.3 & 86.5 &  24.3 & 84.3 & 28.3 & 84.2 \\
        Aya-23-35B &  42.8 & 84.6 & 30.7 & 80.7 & 27.5 & 84.7 & 20.6 & 86.4 &  24.9 & 84.0 & 29.3 & 84.1 \\
        Aya-101 & 25.4 & 78.6 &   25.1 &  75.1  & 22.1 & 83.1 &   14.1 & 84.6 &  19.7 & 82.7 &  21.3 & 80.8  \\
        LLaMAX3-Alpaca-8B &  34.0 & 81.5 &   20.9 & 73.3 & 23.5 & 81.6 &  11.9 & 81.8 & 19.8 & 80.6 & 22.0 & 79.8\\
        X-ALMA-13B (Only SFT, MoE) & 47.5 & 86.1 & 40.9 & 84.1  &  31.5 & 85.9  &  22.3  & 86.8 &  27.4  & 85.3 &  33.9 & 85.6 \\
        \midrule
        \textbf{X-ALMA-13B-DAT (MoE)} & 40.3 & 85.0 & 35.3 & 83.3 & 27.6 & 85.0 & 19.0 & 86.0 & 23.5 & 84.6 & 29.1 & 84.8 \\
        \textbf{X-ALMA-13B-DATM (MoE)} & 40.3 & 85.0 & 35.3 & 83.3 & 27.6 & 85.0 & 19.0 & 86.0 & 23.5 & 84.6 & 29.1 & 84.8 \\
        \bottomrule
    \end{tabular}}
    \caption{WMT23 En$\rightarrow$XX translation results (BLEU and COMET). The results of baselines are directly sourced from \citet{xu2024x}. We keep the same generation configuration as \citet{xu2024x}.}
    \label{tab:wmt_en_xx}
\end{table*}

\begin{table*}[h]
    \centering
    \resizebox{\textwidth}{!}{
    \begin{tabular}{lcccccccc}
        \toprule
        Direction & NLLB-3.3B & LLaMAX3-Alpaca-8B & Aya-101 & Aya-23-8B & Aya-23-35B & X-ALMA-13B (only SFT) & X-ALMA-13B-DAT (Ours) & X-ALMA-13B-DATM (Ours) \\
        \midrule
        \multicolumn{9}{c}{\textbf{Group 1 (Af, Da, Nl, De, Is, No, Sv)}} \\
        \midrule
        \texttt{en}$\rightarrow$\texttt{af} & 87.4 / 38.9 & 86.0 / 38.5 & 78.8 / 22.5 & 79.6 / 17.6 & 81.2 / 26.7 & 87.5 / 44.2 & \multicolumn{2}{c}{86.6 / 40.7}   \\
        \texttt{en}$\rightarrow$\texttt{da} & 90.0 / 44.5 & 88.6 / 38.2 & 87.6 / 34.2 & 76.4 / 19.3 & 82.9 / 29.0 & 91.8 / 48.6 &  \multicolumn{2}{c}{91.1 / 42.8}   \\
        \texttt{en}$\rightarrow$\texttt{de} & 88.1 / 40.0 & 85.4 / 31.4 & 84.3 / 29.3 & 88.1 / 36.8 & 88.1 / 37.0 & 88.7 / 41.2 &  \multicolumn{2}{c}{88.0 / 36.1}  \\
        \texttt{en}$\rightarrow$\texttt{is} & 84.6 / 24.5 & 81.2 / 18.3 & 84.3 / 20.9 & 38.4 / 1.6 & 51.0 / 5.9 & 87.2 / 28.0 &  \multicolumn{2}{c}{86.8 / 24.2}  \\
        \texttt{en}$\rightarrow$\texttt{nl} & 87.5 / 27.5 & 86.3 / 23.3 & 85.8 / 22.1 & 87.9 / 26.0 & 87.7 / 26.6 & 88.8 / 29.3 &  \multicolumn{2}{c}{88.2 / 26.4}  \\
        \texttt{en}$\rightarrow$\texttt{no} & 88.9 / 33.0 & 87.8 / 28.0 & 87.5 / 26.9 & 77.3 / 15.7 & 82.4 / 22.1 & 90.6 / 35.0 &  \multicolumn{2}{c}{89.9 / 30.9}  \\
\bottomrule
    \end{tabular}}
    \caption{Full results for Group 1-4 languages on Flores-200 benchmark. The performance of baselines is directly sourced from \citet{xu2024x} and we keep the generation configuration of our approach the same as those.}
    \label{tab:translation_results}
\end{table*}

\begin{table*}[h]
    \centering
    \resizebox{\textwidth}{!}{
    \begin{tabular}{lcccccccc}
        \toprule
        Direction & NLLB-3.3B & LLaMAX3-Alpaca-8B & Aya-101 & Aya-23-8B & Aya-23-35B & X-ALMA-13B (only SFT) & X-ALMA-13B-DAT (Ours) & X-ALMA-13B-DATM (Ours) \\
        \midrule
        \multicolumn{9}{c}{\textbf{Group 5 (Cs, El, Hu, Lt, Lv, Pl)}} \\
        \midrule
        \texttt{en}$\rightarrow$\texttt{cs} & 91.0 / 32.2 & 88.1 / 24.6 & 90.0 / 26.7 & 91.1 / 30.5 & 91.4 / 32.2 & {91.5} / {33.8} &  \multicolumn{2}{c}{91.3 / 30.9}  \\
        \texttt{en}$\rightarrow$\texttt{el} & 89.0 / 27.4 & 86.2 / 20.4 & 86.6 / 21.4 & 89.5 / 26.1 & 89.6 / 27.0 & {89.8} / {27.9} &  \multicolumn{2}{c}{89.0 / 23.6}  \\
        \texttt{en}$\rightarrow$\texttt{hu} & 89.3 / 26.4 & 86.6 / 18.2 & 88.4 / 21.4 & 51.7 / 3.6  & 77.0 / 10.8 & {90.4} / {27.0} &  \multicolumn{2}{c}{89.6 / 21.9}  \\
        \texttt{en}$\rightarrow$\texttt{lt} & 89.3 / 25.2 & 86.1 / 17.0 & 89.2 / 22.5 & 65.4 / 5.4  & 82.5 / 14.0 & {91.3} / {28.4} &  \multicolumn{2}{c}{90.0 / 22.5}  \\
        \texttt{en}$\rightarrow$\texttt{lv} & 87.4 / 25.0 & 85.8 / 21.1 & 88.6 / 25.0 & 36.5 / 1.5  & 62.7 / 7.9  & {90.7} / {29.3} &  \multicolumn{2}{c}{89.8 / 24.6}  \\
        \texttt{en}$\rightarrow$\texttt{pl} & 88.9 / 21.6 & 86.7 / 17.2 & 87.6 / 18.3 & 89.2 / 20.7 & 89.8 / 22.4 & {90.1} / {23.3} &  \multicolumn{2}{c}{89.7 / 20.3}  \\
\bottomrule
    \end{tabular}}
    % \caption{Translation Results for Group 5: COMET-22 / BLEU}
    \caption{Full results for Group 5-8 languages on Flores-200 benchmark. The performance of baselines is directly sourced from \citet{xu2024x} and we keep the generation configuration of our approach the same as those.}
    \label{tab:translation_results_group5}
\end{table*}


\begin{thebibliography}{38}
\providecommand{\natexlab}[1]{#1}

\bibitem[{Aharoni et~al.(2019)Aharoni, Johnson, and Firat}]{aharoni-etal-2019-massively}
Roee Aharoni, Melvin Johnson, and Orhan Firat. 2019.
\newblock \href {https://doi.org/10.18653/v1/N19-1388} {Massively multilingual neural machine translation}.
\newblock In \emph{Proceedings of the 2019 Conference of the North {A}merican Chapter of the Association for Computational Linguistics: Human Language Technologies, Volume 1 (Long and Short Papers)}, pages 3874--3884, Minneapolis, Minnesota. Association for Computational Linguistics.

\bibitem[{Alves et~al.(2024)Alves, Pombal, Guerreiro, Martins, Alves, Farajian, Peters, Rei, Fernandes, Agrawal et~al.}]{alves2024tower}
Duarte~M Alves, Jos{\'e} Pombal, Nuno~M Guerreiro, Pedro~H Martins, Jo{\~a}o Alves, Amin Farajian, Ben Peters, Ricardo Rei, Patrick Fernandes, Sweta Agrawal, et~al. 2024.
\newblock Tower: An open multilingual large language model for translation-related tasks.
\newblock \emph{arXiv preprint arXiv:2402.17733}.

\bibitem[{Arivazhagan et~al.(2019)Arivazhagan, Bapna, Firat, Lepikhin, Johnson, Krikun, Chen, Cao, Foster, Cherry et~al.}]{arivazhagan2019massively}
Naveen Arivazhagan, Ankur Bapna, Orhan Firat, Dmitry Lepikhin, Melvin Johnson, Maxim Krikun, Mia~Xu Chen, Yuan Cao, George Foster, Colin Cherry, et~al. 2019.
\newblock Massively multilingual neural machine translation in the wild: Findings and challenges.
\newblock \emph{arXiv preprint arXiv:1907.05019}.

\bibitem[{Aryabumi et~al.(2024)Aryabumi, Dang, Talupuru, Dash, Cairuz, Lin, Venkitesh, Smith, Campos, Tan et~al.}]{aryabumi2024aya}
Viraat Aryabumi, John Dang, Dwarak Talupuru, Saurabh Dash, David Cairuz, Hangyu Lin, Bharat Venkitesh, Madeline Smith, Jon~Ander Campos, Yi~Chern Tan, et~al. 2024.
\newblock Aya 23: Open weight releases to further multilingual progress.
\newblock \emph{arXiv preprint arXiv:2405.15032}.

\bibitem[{Barrault et~al.(2019)Barrault, Bojar, Costa-Jussa, Federmann, Fishel, Graham, Haddow, Huck, Koehn, Malmasi et~al.}]{barrault2019findings}
Lo{\"\i}c Barrault, Ond{\v{r}}ej Bojar, Marta~R Costa-Jussa, Christian Federmann, Mark Fishel, Yvette Graham, Barry Haddow, Matthias Huck, Philipp Koehn, Shervin Malmasi, et~al. 2019.
\newblock Findings of the 2019 conference on machine translation (wmt19).
\newblock ACL.

\bibitem[{Brown et~al.(2020)Brown, Mann, Ryder, Subbiah, Kaplan, Dhariwal, Neelakantan, Shyam, Sastry, Askell et~al.}]{brown2020language}
Tom Brown, Benjamin Mann, Nick Ryder, Melanie Subbiah, Jared~D Kaplan, Prafulla Dhariwal, Arvind Neelakantan, Pranav Shyam, Girish Sastry, Amanda Askell, et~al. 2020.
\newblock Language models are few-shot learners.
\newblock \emph{Advances in neural information processing systems}, 33:1877--1901.

\bibitem[{Conneau(2019)}]{conneau2019unsupervised}
A~Conneau. 2019.
\newblock Unsupervised cross-lingual representation learning at scale.
\newblock \emph{arXiv preprint arXiv:1911.02116}.

\bibitem[{Cui et~al.(2025)Cui, Gao, Liu, Luan et~al.}]{cui2025multilingual}
Menglong Cui, Pengzhi Gao, Wei Liu, Jian Luan, et~al. 2025.
\newblock Multilingual machine translation with open large language models at practical scale: An empirical study.
\newblock \emph{arXiv preprint arXiv:2502.02481}.

\bibitem[{Dang et~al.(2024)Dang, Singh, D'souza, Ahmadian, Salamanca, Smith, Peppin, Hong, Govindassamy, Zhao et~al.}]{dang2024aya}
John Dang, Shivalika Singh, Daniel D'souza, Arash Ahmadian, Alejandro Salamanca, Madeline Smith, Aidan Peppin, Sungjin Hong, Manoj Govindassamy, Terrence Zhao, et~al. 2024.
\newblock Aya expanse: Combining research breakthroughs for a new multilingual frontier.
\newblock \emph{arXiv preprint arXiv:2412.04261}.

\bibitem[{Dubey et~al.(2024)Dubey, Jauhri, Pandey, Kadian, Al-Dahle, Letman, Mathur, Schelten, Yang, Fan et~al.}]{dubey2024llama}
Abhimanyu Dubey, Abhinav Jauhri, Abhinav Pandey, Abhishek Kadian, Ahmad Al-Dahle, Aiesha Letman, Akhil Mathur, Alan Schelten, Amy Yang, Angela Fan, et~al. 2024.
\newblock The llama 3 herd of models.
\newblock \emph{arXiv preprint arXiv:2407.21783}.

\bibitem[{Fan et~al.(2021)Fan, Bhosale, Schwenk, Ma, El-Kishky, Goyal, Baines, Celebi, Wenzek, Chaudhary et~al.}]{fan2021beyond}
Angela Fan, Shruti Bhosale, Holger Schwenk, Zhiyi Ma, Ahmed El-Kishky, Siddharth Goyal, Mandeep Baines, Onur Celebi, Guillaume Wenzek, Vishrav Chaudhary, et~al. 2021.
\newblock Beyond english-centric multilingual machine translation.
\newblock \emph{Journal of Machine Learning Research}, 22(107):1--48.

\bibitem[{Guo et~al.(2024)Guo, Yang, Li, Wei, Shang, and Chen}]{guo-etal-2024-novel}
Jiaxin Guo, Hao Yang, Zongyao Li, Daimeng Wei, Hengchao Shang, and Xiaoyu Chen. 2024.
\newblock \href {https://doi.org/10.18653/v1/2024.findings-naacl.42} {A novel paradigm boosting translation capabilities of large language models}.
\newblock In \emph{Findings of the Association for Computational Linguistics: NAACL 2024}, pages 639--649, Mexico City, Mexico. Association for Computational Linguistics.

\bibitem[{Han et~al.(2024)Han, Eriguchi, Xu, Hoang, Carpuat, and Khayrallah}]{han2024adapters}
HyoJung Han, Akiko Eriguchi, Haoran Xu, Hieu Hoang, Marine Carpuat, and Huda Khayrallah. 2024.
\newblock Adapters for altering llm vocabularies: What languages benefit the most?
\newblock \emph{arXiv preprint arXiv:2410.09644}.

\bibitem[{He et~al.(2024)He, Liang, Jiao, Zhang, Yang, Wang, Tu, Shi, and Wang}]{he-etal-2024-exploring}
Zhiwei He, Tian Liang, Wenxiang Jiao, Zhuosheng Zhang, Yujiu Yang, Rui Wang, Zhaopeng Tu, Shuming Shi, and Xing Wang. 2024.
\newblock \href {https://doi.org/10.1162/tacl_a_00642} {Exploring human-like translation strategy with large language models}.
\newblock \emph{Transactions of the Association for Computational Linguistics}, 12:229--246.

\bibitem[{Hu et~al.(2022)Hu, yelong shen, Wallis, Allen-Zhu, Li, Wang, Wang, and Chen}]{hu2022lora}
Edward~J Hu, yelong shen, Phillip Wallis, Zeyuan Allen-Zhu, Yuanzhi Li, Shean Wang, Lu~Wang, and Weizhu Chen. 2022.
\newblock \href {https://openreview.net/forum?id=nZeVKeeFYf9} {Lo{RA}: Low-rank adaptation of large language models}.
\newblock In \emph{International Conference on Learning Representations}.

\bibitem[{Lin et~al.(2019)Lin, Chen, Lee, Li, Zhang, Xia, Rijhwani, He, Zhang, Ma, Anastasopoulos, Littell, and Neubig}]{lin-etal-2019-choosing}
Yu-Hsiang Lin, Chian-Yu Chen, Jean Lee, Zirui Li, Yuyan Zhang, Mengzhou Xia, Shruti Rijhwani, Junxian He, Zhisong Zhang, Xuezhe Ma, Antonios Anastasopoulos, Patrick Littell, and Graham Neubig. 2019.
\newblock \href {https://doi.org/10.18653/v1/P19-1301} {Choosing transfer languages for cross-lingual learning}.
\newblock In \emph{Proceedings of the 57th Annual Meeting of the Association for Computational Linguistics}, pages 3125--3135, Florence, Italy. Association for Computational Linguistics.

\bibitem[{Liu et~al.(2023)Liu, Huang, Yu, Li, Su, and Huang}]{liu-etal-2023-continual}
Junpeng Liu, Kaiyu Huang, Hao Yu, Jiuyi Li, Jinsong Su, and Degen Huang. 2023.
\newblock \href {https://doi.org/10.18653/v1/2023.emnlp-main.736} {Continual learning for multilingual neural machine translation via dual importance-based model division}.
\newblock In \emph{Proceedings of the 2023 Conference on Empirical Methods in Natural Language Processing}, pages 12011--12027, Singapore. Association for Computational Linguistics.

\bibitem[{Lu et~al.(2024{\natexlab{a}})Lu, Yang, Huang, Zhang, Lam, and Wei}]{lu-etal-2024-chain}
Hongyuan Lu, Haoran Yang, Haoyang Huang, Dongdong Zhang, Wai Lam, and Furu Wei. 2024{\natexlab{a}}.
\newblock \href {https://doi.org/10.18653/v1/2024.emnlp-main.55} {Chain-of-dictionary prompting elicits translation in large language models}.
\newblock In \emph{Proceedings of the 2024 Conference on Empirical Methods in Natural Language Processing}, pages 958--976, Miami, Florida, USA. Association for Computational Linguistics.

\bibitem[{Lu et~al.(2024{\natexlab{b}})Lu, Zhu, Li, Qiao, and Yuan}]{lu2024llamax}
Yinquan Lu, Wenhao Zhu, Lei Li, Yu~Qiao, and Fei Yuan. 2024{\natexlab{b}}.
\newblock Llamax: Scaling linguistic horizons of llm by enhancing translation capabilities beyond 100 languages.
\newblock \emph{arXiv preprint arXiv:2407.05975}.

\bibitem[{Meng and Monz(2024)}]{meng-monz-2024-disentangling}
Yan Meng and Christof Monz. 2024.
\newblock \href {https://aclanthology.org/2024.eacl-long.110/} {Disentangling the roles of target-side transfer and regularization in multilingual machine translation}.
\newblock In \emph{Proceedings of the 18th Conference of the European Chapter of the Association for Computational Linguistics (Volume 1: Long Papers)}, pages 1828--1840, St. Julian{'}s, Malta. Association for Computational Linguistics.

\bibitem[{Post(2018)}]{post-2018-call}
Matt Post. 2018.
\newblock \href {https://doi.org/10.18653/v1/W18-6319} {A call for clarity in reporting {BLEU} scores}.
\newblock In \emph{Proceedings of the Third Conference on Machine Translation: Research Papers}, pages 186--191, Brussels, Belgium. Association for Computational Linguistics.

\bibitem[{Rei et~al.(2022)Rei, C.~de Souza, Alves, Zerva, Farinha, Glushkova, Lavie, Coheur, and Martins}]{rei-etal-2022-comet}
Ricardo Rei, Jos{\'e}~G. C.~de Souza, Duarte Alves, Chrysoula Zerva, Ana~C Farinha, Taisiya Glushkova, Alon Lavie, Luisa Coheur, and Andr{\'e} F.~T. Martins. 2022.
\newblock \href {https://aclanthology.org/2022.wmt-1.52/} {{COMET}-22: Unbabel-{IST} 2022 submission for the metrics shared task}.
\newblock In \emph{Proceedings of the Seventh Conference on Machine Translation (WMT)}, pages 578--585, Abu Dhabi, United Arab Emirates (Hybrid). Association for Computational Linguistics.

\bibitem[{Shaham et~al.(2023)Shaham, Elbayad, Goswami, Levy, and Bhosale}]{shaham-etal-2023-causes}
Uri Shaham, Maha Elbayad, Vedanuj Goswami, Omer Levy, and Shruti Bhosale. 2023.
\newblock \href {https://doi.org/10.18653/v1/2023.acl-long.883} {Causes and cures for interference in multilingual translation}.
\newblock In \emph{Proceedings of the 61st Annual Meeting of the Association for Computational Linguistics (Volume 1: Long Papers)}, pages 15849--15863, Toronto, Canada. Association for Computational Linguistics.

\bibitem[{Tang et~al.(2020)Tang, Tran, Li, Chen, Goyal, Chaudhary, Gu, and Fan}]{tang2020multilingual}
Yuqing Tang, Chau Tran, Xian Li, Peng-Jen Chen, Naman Goyal, Vishrav Chaudhary, Jiatao Gu, and Angela Fan. 2020.
\newblock Multilingual translation with extensible multilingual pretraining and finetuning.
\newblock \emph{arXiv preprint arXiv:2008.00401}.

\bibitem[{{\"U}st{\"u}n et~al.(2024){\"U}st{\"u}n, Aryabumi, Yong, Ko, D{'}souza, Onilude, Bhandari, Singh, Ooi, Kayid, Vargus, Blunsom, Longpre, Muennighoff, Fadaee, Kreutzer, and Hooker}]{ustun-etal-2024-aya}
Ahmet {\"U}st{\"u}n, Viraat Aryabumi, Zheng Yong, Wei-Yin Ko, Daniel D{'}souza, Gbemileke Onilude, Neel Bhandari, Shivalika Singh, Hui-Lee Ooi, Amr Kayid, Freddie Vargus, Phil Blunsom, Shayne Longpre, Niklas Muennighoff, Marzieh Fadaee, Julia Kreutzer, and Sara Hooker. 2024.
\newblock \href {https://doi.org/10.18653/v1/2024.acl-long.845} {Aya model: An instruction finetuned open-access multilingual language model}.
\newblock In \emph{Proceedings of the 62nd Annual Meeting of the Association for Computational Linguistics (Volume 1: Long Papers)}, pages 15894--15939, Bangkok, Thailand. Association for Computational Linguistics.

\bibitem[{Wang and Neubig(2019)}]{wang-neubig-2019-target}
Xinyi Wang and Graham Neubig. 2019.
\newblock \href {https://doi.org/10.18653/v1/P19-1583} {Target conditioned sampling: Optimizing data selection for multilingual neural machine translation}.
\newblock In \emph{Proceedings of the 57th Annual Meeting of the Association for Computational Linguistics}, pages 5823--5828, Florence, Italy. Association for Computational Linguistics.

\bibitem[{Wang et~al.(2020)Wang, Tsvetkov, and Neubig}]{wang-etal-2020-balancing}
Xinyi Wang, Yulia Tsvetkov, and Graham Neubig. 2020.
\newblock \href {https://doi.org/10.18653/v1/2020.acl-main.754} {Balancing training for multilingual neural machine translation}.
\newblock In \emph{Proceedings of the 58th Annual Meeting of the Association for Computational Linguistics}, pages 8526--8537, Online. Association for Computational Linguistics.

\bibitem[{Wei et~al.(2022)Wei, Wang, Schuurmans, Bosma, Xia, Chi, Le, Zhou et~al.}]{wei2022chain}
Jason Wei, Xuezhi Wang, Dale Schuurmans, Maarten Bosma, Fei Xia, Ed~Chi, Quoc~V Le, Denny Zhou, et~al. 2022.
\newblock Chain-of-thought prompting elicits reasoning in large language models.
\newblock \emph{Advances in neural information processing systems}, 35:24824--24837.

\bibitem[{Xu et~al.(2024{\natexlab{a}})Xu, Kim, Sharaf, and Awadalla}]{xu2024a}
Haoran Xu, Young~Jin Kim, Amr Sharaf, and Hany~Hassan Awadalla. 2024{\natexlab{a}}.
\newblock \href {https://openreview.net/forum?id=farT6XXntP} {A paradigm shift in machine translation: Boosting translation performance of large language models}.
\newblock In \emph{The Twelfth International Conference on Learning Representations}.

\bibitem[{Xu et~al.(2024{\natexlab{b}})Xu, Murray, Koehn, Hoang, Eriguchi, and Khayrallah}]{xu2024x}
Haoran Xu, Kenton Murray, Philipp Koehn, Hieu Hoang, Akiko Eriguchi, and Huda Khayrallah. 2024{\natexlab{b}}.
\newblock X-alma: Plug \& play modules and adaptive rejection for quality translation at scale.
\newblock \emph{arXiv preprint arXiv:2410.03115}.

\bibitem[{Xu et~al.(2024{\natexlab{c}})Xu, Sharaf, Chen, Tan, Shen, Van~Durme, Murray, and Kim}]{xu2024contrastive}
Haoran Xu, Amr Sharaf, Yunmo Chen, Weiting Tan, Lingfeng Shen, Benjamin Van~Durme, Kenton Murray, and Young~Jin Kim. 2024{\natexlab{c}}.
\newblock Contrastive preference optimization: Pushing the boundaries of llm performance in machine translation.
\newblock \emph{arXiv preprint arXiv:2401.08417}.

\bibitem[{Xu et~al.(2023)Xu, Tan, Li, Chen, Van~Durme, Koehn, and Murray}]{xu-etal-2023-condensing}
Haoran Xu, Weiting Tan, Shuyue Li, Yunmo Chen, Benjamin Van~Durme, Philipp Koehn, and Kenton Murray. 2023.
\newblock \href {https://doi.org/10.18653/v1/2023.emnlp-main.97} {Condensing multilingual knowledge with lightweight language-specific modules}.
\newblock In \emph{Proceedings of the 2023 Conference on Empirical Methods in Natural Language Processing}, pages 1575--1587, Singapore. Association for Computational Linguistics.

\bibitem[{Yadav et~al.(2024)Yadav, Tam, Choshen, Raffel, and Bansal}]{yadav2024ties}
Prateek Yadav, Derek Tam, Leshem Choshen, Colin~A Raffel, and Mohit Bansal. 2024.
\newblock Ties-merging: Resolving interference when merging models.
\newblock \emph{Advances in Neural Information Processing Systems}, 36.

\bibitem[{Yang et~al.(2023)Yang, Li, Zhang, and Zong}]{yang2023bigtranslate}
Wen Yang, Chong Li, Jiajun Zhang, and Chengqing Zong. 2023.
\newblock Bigtranslate: Augmenting large language models with multilingual translation capability over 100 languages.
\newblock \emph{arXiv preprint arXiv:2305.18098}.

\bibitem[{Yu et~al.(2024)Yu, Yu, Yu, Huang, and Li}]{yu2024language}
Le~Yu, Bowen Yu, Haiyang Yu, Fei Huang, and Yongbin Li. 2024.
\newblock Language models are super mario: Absorbing abilities from homologous models as a free lunch.
\newblock In \emph{Forty-first International Conference on Machine Learning}.

\bibitem[{Zhang et~al.(2023)Zhang, Liu, He et~al.}]{zhang2023composing}
Jinghan Zhang, Junteng Liu, Junxian He, et~al. 2023.
\newblock Composing parameter-efficient modules with arithmetic operation.
\newblock \emph{Advances in Neural Information Processing Systems}, 36:12589--12610.

\bibitem[{Zhao et~al.(2024)Zhao, Chen, Cheng, and Chen}]{zhao2024sparse}
Xinyu Zhao, Xuxi Chen, Yu~Cheng, and Tianlong Chen. 2024.
\newblock \href {https://openreview.net/forum?id=ySS7hH1smL} {Sparse moe with language guided routing for multilingual machine translation}.
\newblock In \emph{The Twelfth International Conference on Learning Representations}.

\bibitem[{Zheng et~al.(2024)Zheng, Li, Bao, Wang, Shan, Xiao, and Zhu}]{zheng-etal-2024-partialformer}
Tong Zheng, Bei Li, Huiwen Bao, Jiale Wang, Weiqiao Shan, Tong Xiao, and JingBo Zhu. 2024.
\newblock \href {https://doi.org/10.18653/v1/2024.findings-acl.434} {{P}artial{F}ormer: Modeling part instead of whole for machine translation}.
\newblock In \emph{Findings of the Association for Computational Linguistics: ACL 2024}, pages 7280--7294, Bangkok, Thailand. Association for Computational Linguistics.

\end{thebibliography}
\end{document}